\documentclass[journal]{IEEEtran}

\usepackage[linesnumbered,ruled]{algorithm2e}
\usepackage{graphicx,amssymb,mathrsfs,amsmath}
\usepackage{epstopdf}
\usepackage{subfigure}
\usepackage{epsfig}
\usepackage{setspace}
\usepackage{multirow}
\usepackage{booktabs}
\usepackage{soul}

\newcommand{\norm}[1]{\lVert#1\rVert}
\usepackage{algpseudocode}
\usepackage{cite}

\makeatletter
\def\hlinew#1{%
  \noalign{\ifnum0=`}\fi\hrule \@height #1 \futurelet
   \reserved@a\@xhline}
\DeclareMathOperator{\tr}{tr}
\DeclareMathOperator{\F}{F}
\DeclareMathOperator{\T}{T}
\DeclareMathOperator{\svd}{svd}
\newcommand{\maxIter}{\textit{maxIter}}

\begin{document}
\setstcolor{red}
\title{CoSpace: Common Subspace Learning from Hyperspectral-Multispectral Correspondences}

\author{Danfeng Hong,~\IEEEmembership{Student Member,~IEEE,}
        Naoto Yokoya,~\IEEEmembership{Member,~IEEE,}
        Jocelyn Chanussot,~\IEEEmembership{Fellow,~IEEE,}
        and~Xiao Xiang Zhu,~\IEEEmembership{Senior Member,~IEEE}
\thanks{This work was supported by funding from the European Research Council (ERC) under the European Union's Horizon 2020 research and innovation programme (grant agreement No [ERC-2016-StG-714087]) and from Helmholtz Association under the framework of the Young Investigators Group ''SiPEO'' (VH-NG-1018, www.sipeo.bgu.tum.de). The work of N. Yokoya was supported by Japan Society for the Promotion of Science (JSPS) KAKENHI 15K20955 and Alexander von Humboldt Fellowship for postdoctoral researchers. (\emph{Corresponding author: Xiao Xiang Zhu.}) }
\thanks{D. Hong and X. Zhu are with the Remote Sensing Technology Institute (IMF), German Aerospace Center (DLR), 82234 Wessling, Germany, and Signal Processing in Earth Observation (SiPEO), Technical University of Munich (TUM), 80333 Munich, Germany. (e-mail: danfeng.hong@dlr.de; xiao.zhu@dlr.de)}
\thanks{N. Yokoya is with the Geoinformatics Unit, RIKEN Center for Advanced Intelligence Project (AIP), RIKEN, 103-0027 Tokyo, Japan. (e-mail: naoto.yokoya@riken.jp)}
\thanks{J. Chanussot is with the Univ. Grenoble Alpes, CNRS, Grenoble INP, GIPSA-lab, F-38000 Grenoble, France, also with the Faculty of Electrical and Computer Engineering, University of Iceland, Reykjavik 101, Iceland. (e-mail: jocelyn@hi.is)}
}

\markboth{IEEE Transactions on Geoscience and Remote Sensing,~Vol.~XX, No.~XX, ~XXXX,~2018}%
{Shell \MakeLowercase{\textit{et al.}}: CoSpace: Common Subspace Learning from Hyperspectral-Multispectral Correspondences}
\maketitle
\begin{abstract}
With a large amount of open satellite multispectral imagery (e.g., Sentinel-2 and Landsat-8), considerable attention has been paid to global multispectral land cover classification. However, its limited spectral information hinders further improving the classification performance. Hyperspectral imaging enables discrimination between spectrally similar classes but its swath width from space is narrow compared to multispectral ones. To achieve accurate land cover classification over a large coverage, we propose a cross-modality feature learning framework, called common subspace learning (CoSpace), by jointly considering subspace learning and supervised classification. By locally aligning the manifold structure of the two modalities, CoSpace linearly learns a shared latent subspace from hyperspectral-multispectral (HS-MS) correspondences. The multispectral out-of-samples can be then projected into the subspace, which are expected to take advantages of rich spectral information of the corresponding hyperspectral data used for learning, and thus leads to a better classification. Extensive experiments on two simulated HS-MS datasets (University of Houston and Chikusei), where HS-MS data sets have trade-offs between coverage and spectral resolution, are performed to demonstrate the superiority and effectiveness of the proposed method in comparison with previous state-of-the-art methods.
\end{abstract}
\graphicspath{{figures/}}
\section{Introduction}
\IEEEPARstart{R}{ecently}, the launch of operational optical broadband (multispectral) satellites has successfully boosted the usage of multispectral data for various tasks such as urban monitoring, management of natural resources, ecosystem, and disasters prediction. There has been a growing interest in large-scale land cover mapping of urban \cite{du2012fusion}, agriculture monitoring \cite{yang2013Agriculture}\cite{xie2016dynamic}, and mineral exploration \cite{van2014potential}, since high-quality multispectral (MS) satellite imagery is openly available on a global scale (e.g., Sentinel-2 and Landsat-8). However, MS data fails to discriminate spectrally similar classes due to its broad spectral bandwidth. Hyperspectral imaging can acquire richer spectral information that enables high discrimination ability but its coverage from space is much narrower than the one of MS imaging due to the limitations of imaging devices and satellite techniques. This trade-off naturally motivates us to ponder a question: \emph{can HS imagery covering only a limited part of the MS imagery be explored to improve the classification of the entire area covered by the MS imagery?} This is as a typical cross-modal feature learning problem.

\begin{figure}[t]
	  \centering
		\subfigure{
			\includegraphics[width=0.49\textwidth]{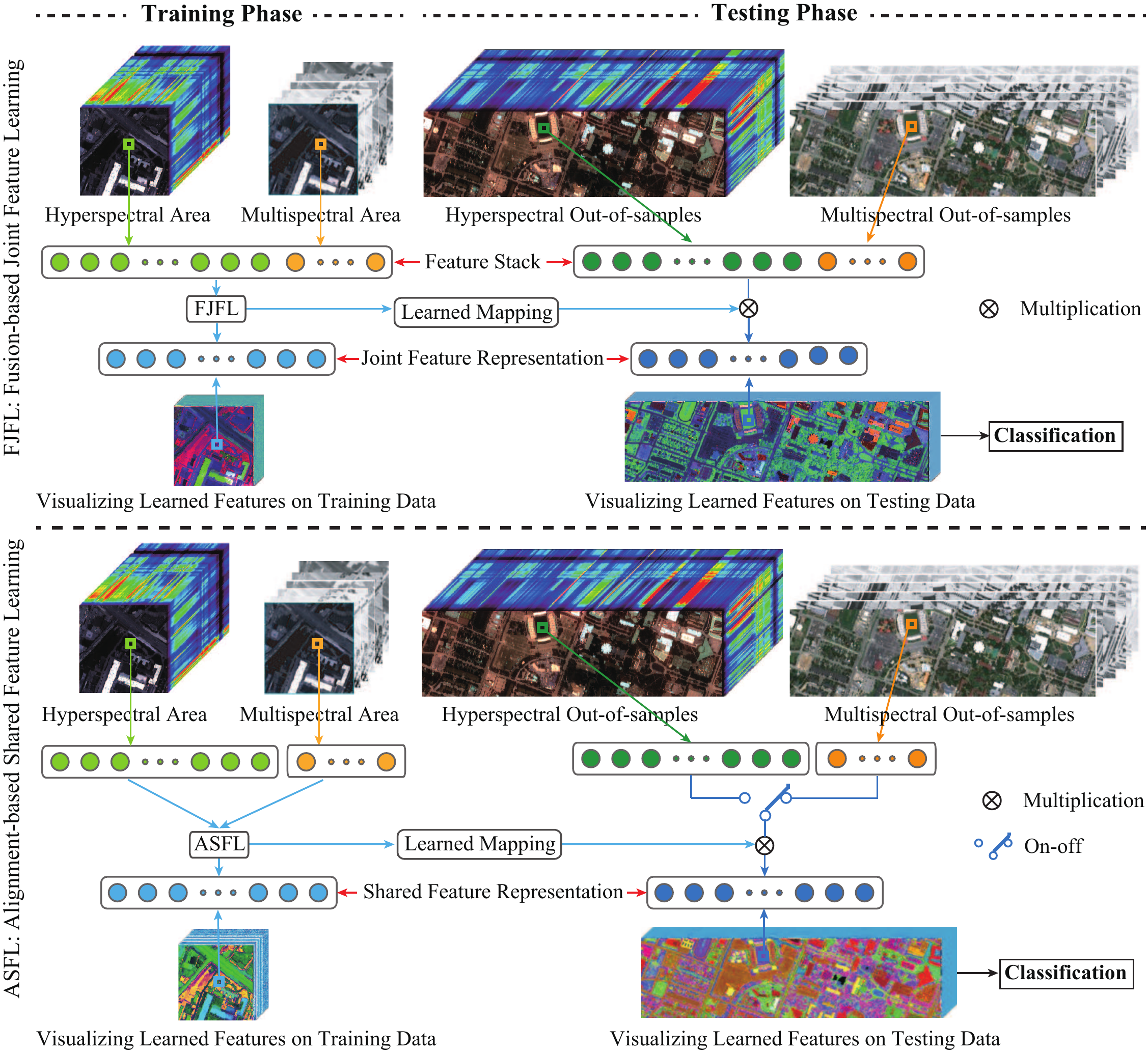}
		}
        \caption{An illustration of the two kinds of multi-modal feature learning frameworks, where the switch (On-off) means that only one modality is involved as the testing samples to meet the hypothesis of cross-modal learning.}
\label{fig:ClarifyCrossModality}
\end{figure}

Researchers have proposed a variety of multi-modal feature learning algorithms by introducing additional information, which can be roughly categorized into two parts: 
fusion-based joint feature learning (FJFL) \cite{yokoya2017hyperspectral,chen2017FeatureStack} and alignment-based shared feature learning (ASFL) \cite{tuia2014ManifoldAlignment}. The main difference between FJFL and ASFL is illustrated in Fig. \ref{fig:ClarifyCrossModality}.

FJFL aims to learn discriminative features by absorbing the different properties from multi-modal data. FJFL fuses the different sources at the data level to diversify the information and then to further learn the higher-level feature representation. One intuitive way for FJFL is to directly learn a joint data representation at the feature level. At present, this is the mainstream approach for multi-modal data analysis \cite{yokoya2012coupled}. For example, by embedding the height information from LiDAR into MS (HS) data, Ghamisi \emph{et al.} \cite{ghamisi2017hyperspectral} learned multi-fold features from HS and LiDAR correspondences for a multi-modal classification task. Iyer \emph{et al.} \cite{iyer2017graph} provided a graph-based new perspective for feature extraction and segmentation of multi-modal images and achieved a desirable result. The resulting discriminative features are beneficial for improving the performance of some high-level applications, especially classification \cite{hong2016local}\cite{hong2017DR}, object detection \cite{liu2017multiscale}, image/video analysis \cite{tochon2017Video}, and spectral unmixing \cite{hong2017unmixing}. Image fusion can be also regarded as a part of FJFL when feature learning is applied subsequently. For instance, hyperspectral and multispectral (HS-MS) data fusion enhances the spectral resolution of MS data by fusing it with the low-spatial-resolution HS data \cite{yokoya2017hyperspectral}. The fused HS-MS product can be then seen as a new input for further discriminative feature learning.  

\begin{figure}[!t]
	  \centering
		\subfigure{
			\includegraphics[width=0.48\textwidth]{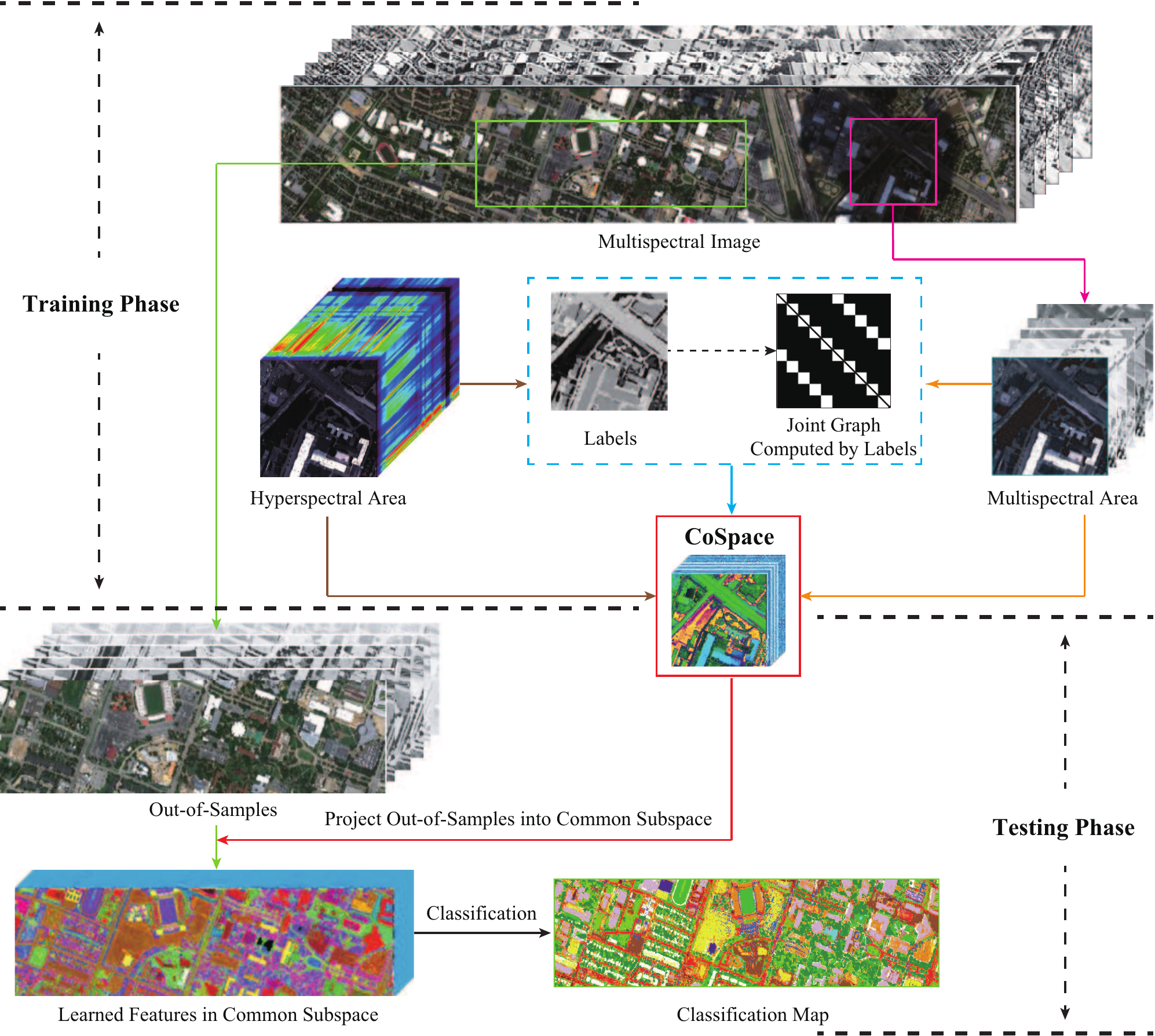}
		}
        \caption{The holistic workflow of the proposed CoSpace.}
\label{fig:CoSpaceframework}
\end{figure}

Behind the advancement of FJFL, the \emph{complete data correspondence} is the prerequisite. This limitation undoubtedly results in a poor fit for cross-modal data analysis, in particular for cross-modal feature learning \cite{ngiam2011multimodal}\footnote{In contrast to multi-modal learning (bi-modality for example), cross-modal learning trains on single modality and tests on bi-modality, or \emph{vice versa }(train on bi-modality and test on single modality).}. In our MS-HS case, the cross-modal learning refers to a problem that given a large-scale MS image and a limited HS area partially overlapping with the MS data (see Fig. \ref{fig:CoSpaceframework} for example), we learn the low-dimensional embedding representation from the limited amount of MS-HS correspondences and transfer the learned features to the rest of MS data for improving the performance of large-scale land-cover and land-use mapping. During the process, we expect to transfer the discrimination capability learnt from the rich spectral information into MS data through the learned common subspace in order to more effectively identify some challenging classes that are hardly recognized by MS data due to its poor spectral information. Please note that we just start a preliminary investigation of cross-modal learning (MS-HS) in this paper, that is, the MS and HS images share the same land-cover classes.

Unlike FJFL, ASFL is more apt for cross-modal feature learning, since ASFL can adaptively shuttle back and forth between the different modalities or domains by means of the learned common subspace. Matasci \emph{et al.} \cite{matasci2011transfer} linearly projected the hyperspectral data of the source and target domains into a common feature space where the gap between domains in hyperspectral image classification are expected to be reduced. Kulis \emph{et al.} \cite{kulis2011you} addressed the issue of visual domain adaption by learning a nonlinear transformation in kernel space, with the application to general object recognition. In \cite{zhou2017domain}, a probabilistic framework was proposed to align the class distributions of two domains for robust hyperspectral image classification. Manifold alignment (MA) \cite{Wang2011MA} is also a powerful tool for modeling this kind of issue. Inspired by MA, Tuia \emph{et al.} \cite{tuia2014ManifoldAlignment} aligned multi-view remote sensing images on manifolds by fully allowing for the spectral variabilities between the different angle imageries, yielding a significant improvement of classification performance. 

It should be noted that these methods mentioned above only consider the differences of a unimodality between the source and target domains at the level of original features, but they fail to investigate the transferability of multi-modality since the different modalities usually hold the different feature dimensions. Although these approaches can build connections between features or instances, a poorly connected relationship between the learned common subspace and label information is still hindering the low-dimensional feature representation from being more discriminative.

We propose a cross-modality feature learning framework, called common subspace learning (CoSpace), that learns the shared feature representation (common subspace) from partial HS-MS correspondences. Extensive experiments are conducted on simulated MS and partially overlapped real HS data based on two airborne HS datasets: the University of Houston and Chikusei datasets. MS data is generated from HS data  by using the spectral response functions (SRFs) of Sentinel-2. We re-label the training and testing classes on the datasets to meet the problem setting of cross-modal feature learning and further to make them more challenging (see Section III for details). Our contributions can be specifically unfolded as follows:
\begin{itemize}
\item We propose a novel common subspace learning (CoSpace) approach by jointly considering the subspace learning and classification in order to effectively bridge the learned features and label information, aiming at addressing the HS-MS cross-modal feature learning issue;
\item By locally aligning HS-MS data on the low-dimensional manifolds where the features of HS and MS share the same dimension, CoSpace linearly learns a latent shared subspace from HS-MS correspondences, where samples are expected to be better classified. Owing to the subspace learned in a linear way, the out-of-samples data can be simply and smoothly embedded;
\item An optimization algorithm based on the alternating direction method of multipliers (ADMM) is designed to solve the proposed model.
\end{itemize}

The remainder of this paper is organized as follows. In section II, we first clarify our motivation and then propose the methodology of the CoSpace model, finally elaborate on the corresponding ADMM-based optimization algorithm. Section III presents the experimental results and analysis on two different HS-MS datasets both qualitatively and quantitatively. Finally, some conclusions are drawn in Section IV.
\section{CoSpace: Common Subspace Learning}
To take the benefit of HS imagery covering only a limited part of the MS imagery, and subsequently improve the classification results of the entire area covered by the MS imagery, our idea is to learn a HS-MS common subspace, in which the data from one domain can be adaptively transferred to another domain.
Fig. \ref{fig:CoSpaceframework} shows the holistic diagram of the proposed CoSpace method.
\begin{algorithm}[!t]
\label{alg1}
\caption{Common Subspace Learning (CoSpace)}
\KwIn{$\widetilde{\mathbf{Y}}$, $\widetilde{\mathbf{X}}$, $\mathbf{L}$, and parameters $\alpha$, $\beta$, $\maxIter$.}
\KwOut{$\mathbf{P}$, $\mathbf{\Theta}$}
 $t=1$, $\zeta=1e-4$;\\
{\textbf{Initializating} $\mathbf{P}$ and $\mathbf{\Theta}$}\\

  \While{not converged \rm{or} $t>\maxIter$}
 {
   Fix other variables to update $\mathbf{P}$ by Eq.~(\ref{eq7})

   Fix other variables to update $\mathbf{\Theta}$ by \textbf{Algorithm} $\mathbf{2}$

   Compute the objective function value $E^{t+1}$ and check the convergence condition:
   \eIf{$|\frac{E^{t+1}-E^{t}}{E^{t}}|<\zeta$}
   {
     Stop iteration;
   }
   {
     $t\leftarrow t+1$;
   }
 }
\end{algorithm}
\subsection{Problem Formulation}
Let $\mathbf{X}_{M} \in\mathbb{R}^{d_{M}\times N}$ and $\mathbf{X}_{H} \in\mathbb{R}^{d_{H}\times N}$ be the observed MS image with $d_{M}$ bands by $N$ pixels and the HS image with $d_{H}$ bands by $N$ pixels, respectively. $\mathbf{Y} \in\mathbb{R}^{L\times N}$ is the label matrix represented by one-hot encoding. $\mathbf{\Theta}_{M} \in\mathbb{R}^{d \times d_{M}}$ ($\mathbf{\Theta}_{H} \in\mathbb{R}^{d \times d_{H}}$) is denoted as the projection matrix for connecting the MS (HS) data and the latent subspace. The variable $\mathbf{P} \in\mathbb{R}^{L \times d}$ is the weighted matrix specified by bridging the latent subspace and label information. Accordingly, $\mathbf{\widetilde{Y}}=\lbrack \mathbf{Y}, \mathbf{Y}\rbrack \in\mathbb{R}^{L\times 2N}$ can be modeled as follows:
\begin{equation}
\label{eq1}
\begin{aligned}
	  \mathbf{\widetilde{Y}}=\mathbf{P}\mathbf{\Theta}\mathbf{\widetilde{X}} + \mathbf{E},
\end{aligned}
\end{equation}
where $\widetilde{\mathbf{X}}=
\begin{bmatrix}
     \mathbf{X}_{M} & \mathbf{0} \\
     \mathbf{0} & \mathbf{X}_{H}
\end{bmatrix} \in\mathbb{R}^{(d_{M}+d_{H})\times 2N}$, and $\mathbf{\Theta}=\lbrack \mathbf{\Theta}_{M}, \mathbf{\Theta}_{H}\rbrack \in\mathbb{R}^{d\times (d_{M}+d_{H})}$. $\mathbf{E} \in\mathbb{R}^{L\times 2N}$ is the corresponding residual matrix containing the additive noise and other errors.
\begin{algorithm}[!t]
\caption{Solving the subproblem for $\mathbf{\Theta}$}
\KwIn{$\widetilde{\mathbf{Y}}$, $\mathbf{P}$, $\mathbf{J}$, $\widetilde{\mathbf{X}}$, $\mathbf{L}$, $\beta$, $\maxIter$.}
\KwOut{$\mathbf{\Theta}.$}
\textbf{Initialization}:
$\mathbf{\Theta}=\mathbf{0}$, $\mathbf{G}=\mathbf{0}$, $\mathbf{\Lambda}_{1}=\mathbf{0}$,
$\mathbf{\Lambda}_{2}=\mathbf{0}$, $\mu=10^{-3}$, $\mu_{\max}=10^{6}$, $\rho=1.5$, $\varepsilon=10^{-6}$, $t=1$.\\
\While{not converged \rm{or} $t>\maxIter$}
 {
         Fix other variables to update $\mathbf{J}$ by
         \begin{equation*}
         \begin{aligned}
                \mathbf{J}=&(\mathbf{P}^{\T}\mathbf{P}+\mu\mathbf{I})^{-1}(\mathbf{P}^{\T}\widetilde{\mathbf{Y}}+\mu\mathbf{\Theta}\widetilde{\mathbf{X}}-\mathbf{\Lambda}_{1}).
         \end{aligned}
         \end{equation*}\\
         Fix other variables to update $\mathbf{\Theta}$ by
         \begin{equation*}
         \begin{aligned}
            \mathbf{\Theta}=&(\mu\mathbf{J}\widetilde{\mathbf{X}}^{\T}+\mathbf{\Lambda}_{1}\widetilde{\mathbf{X}}^{\T}+\mu\mathbf{G}+\mathbf{\Lambda}_{2})\\
            & \times(\mu\widetilde{\mathbf{X}}\widetilde{\mathbf{X}}^{\T}+\mu\mathbf{I}+\beta\widetilde{\mathbf{X}}\mathbf{L}\widetilde{\mathbf{X}}^{\T})^{-1}.
         \end{aligned}
         \end{equation*}\\
         Fix other variables to update $\mathbf{G}$ by
         \begin{equation*}
         \begin{aligned}
                \left[\mathbf{U},\mathbf{S},\mathbf{V}\right]=\svd(\mathbf{\mathbf{\Theta}-\mathbf{\Lambda}_{2}/\mu}), \quad \mathbf{G}=\mathbf{U}\mathbf{I}_{n \times m}\mathbf{V}.
         \end{aligned}
         \end{equation*}\\
         Update Lagrange multipliers by
         \begin{equation*}
         \begin{aligned}
                \mathbf{\Lambda}_{1} \leftarrow \mathbf{\Lambda}_{1}+\mu(\mathbf{J}-\mathbf{\Theta}\widetilde{\mathbf{X}}), \quad \mathbf{\Lambda}_{2} \leftarrow \mathbf{\Lambda}_{2}+\mu(\mathbf{G}-\mathbf{\Theta}).
         \end{aligned}
         \end{equation*}\\
         Update penalty parameter by
         \begin{equation*}
         \begin{aligned}
                \mu=\min (\rho\mu, \mu_{\max}).
         \end{aligned}
         \end{equation*}\\
         Check the convergence conditions:
         \eIf{$\norm {\mathbf{J}-\mathbf{\Theta}\widetilde{\mathbf{X}}}_{\F}<\varepsilon$ and $\norm {\mathbf{G}-\mathbf{\Theta}}_{\F}<\varepsilon$}
         {
           Stop iteration;
         }
         {
         $t\leftarrow t+1$;
         }
 }
\end{algorithm}

Since the Eq. (\ref{eq1}) is a typically ill-posed problem owing to more degrees of flexibility involved (e.g. latent subspace estimation), several assumptions (or prior knowledge) should be introduced into CoSpace using regularization technique. Followed by a popular joint learning framework proposed in \cite{Ji2009JointLearning}, we formulate the CoSpace as the following constrained optimization problem:
\begin{equation}
\label{eq2}
\mathop{\min}_{\mathbf{P},\mathbf{\Theta}}\left\{
\begin{aligned}
\frac{1}{2}\norm{\mathbf{\widetilde{Y}}-&\mathbf{P}\mathbf{\Theta}\mathbf{\widetilde{X}}}_{\F}^{2}+\Phi(\mathbf{P})+\Psi(\mathbf{\Theta})\\
&\mathrm{s.t.} \quad \mathbf{\Theta}\mathbf{\Theta}^{\T}=\mathbf{I}
\end{aligned}
\right\}.
\end{equation}
The two regularization terms in Eq. (\ref{eq2}) are detailed in the following:

To achieve a reliable generalization of our model, the variable $\mathbf{P}$ parameterized by $\alpha$ can be regularized by a Frobenius Norm:
\begin{equation}
\label{eq3}
\begin{aligned}
       \Phi(\mathbf{P})=\frac{\alpha}{2}\norm{\mathbf{P}}_{\F}^{2},
\end{aligned}
\end{equation}
and the prior knowledge with respect to $\mathbf{\Theta}$, resulting in a multi-modal manifold alignment regularization, can be expressed with a joint graph structure as
\begin{equation}
\label{eq4}
\begin{aligned}
       \Psi(\mathbf{\Theta})=\frac{\beta}{2}\tr(\mathbf{\Theta}\mathbf{\widetilde{X}}\mathbf{L}(\mathbf{\Theta}\mathbf{\widetilde{X}})^{\T}),
\end{aligned}
\end{equation}
where $\mathbf{L}=\mathbf{D}-\mathbf{W}\in\mathbb{R}^{2N\times 2N}$ stands for a joint Laplacian matrix, $\mathbf{W}$ that is a corresponding adjacency matrix can be directly inferred from label information in the form of the LDA-like graph \cite{Gu2011IJCAI}:
\begin{equation}
\label{eq5}
  \mathbf{W}_{i,j}=
    \begin{cases}
      \begin{aligned}
      1/N_k, \quad & \text{if \(\mathbf{X}_{i}\) and \(\mathbf{X}_{j}\) belong to the \(k\)-th class;}\\
      0, \quad & \text{otherwise,}
      \end{aligned}
    \end{cases}
\end{equation}
and then $\mathbf{D}$ is computed by $\mathbf{D}_{ii}=\sum_{i\neq j}\mathbf{W}_{i,j}$.
\subsection{Model Optimization}
Considering the nonconvexity of problem (2), an iterative alternating optimization strategy is adopted to solve the convex subproblems of each variable $\mathbf{P}$ and $\mathbf{\Theta}$. An implementation of CoSpace is given in \textbf{Algorithm~1}.

\emph{Optimization with respect to $\mathbf{P}$}: This is a typical least-squares problem with Tikhonov regularization that can be formulated as
\begin{equation}
\label{eq6}
\mathop{\min}_{\mathbf{P}}\left\{
\begin{aligned}
 \frac{1}{2}\norm{\widetilde{\mathbf{Y}}-\mathbf{P}\mathbf{\Theta}\widetilde{\mathbf{X}}}_{\F}^{2}+\frac{\alpha}{2}\norm{\mathbf{P}}_{\F}^{2}
\end{aligned}
\right\},
\end{equation}
which has a closed-form solution
\begin{equation}
\label{eq7}
\begin{aligned}
        \mathbf{P} = (\widetilde{\mathbf{Y}}\mathbf{Q}^{\T})(\mathbf{Q}\mathbf{Q}^{\T}+\alpha\mathbf{I})^{-1},
\end{aligned}
\end{equation}
where $\mathbf{Q}=\mathbf{\Theta}\widetilde{\mathbf{X}}$.

\emph{Optimization with respect to $\mathbf{\Theta}$}: the optimization problem for $\mathbf{\Theta}$ can be formulated as
\begin{equation}
\label{eq8}
 \mathop{\min}_{\mathbf{\Theta}}\left\{
\begin{aligned}
\frac{1}{2}\norm{\widetilde{\mathbf{Y}}-\mathbf{P}&\mathbf{\Theta}\widetilde{\mathbf{X}}}_{\F}^{2}+\frac{\beta}{2}\tr(\mathbf{\Theta}\mathbf{\widetilde{X}}\mathbf{L}(\mathbf{\Theta}\mathbf{\widetilde{X}})^{\T})\\
       &\mathrm{s.t.} \quad \mathbf{\Theta}\mathbf{\Theta}^{\T}=\mathbf{I}
\end{aligned}
\right\}.
\end{equation}
\begin{figure}[!t]
	  \centering
		\subfigure{
			\includegraphics[width=0.3\textwidth]{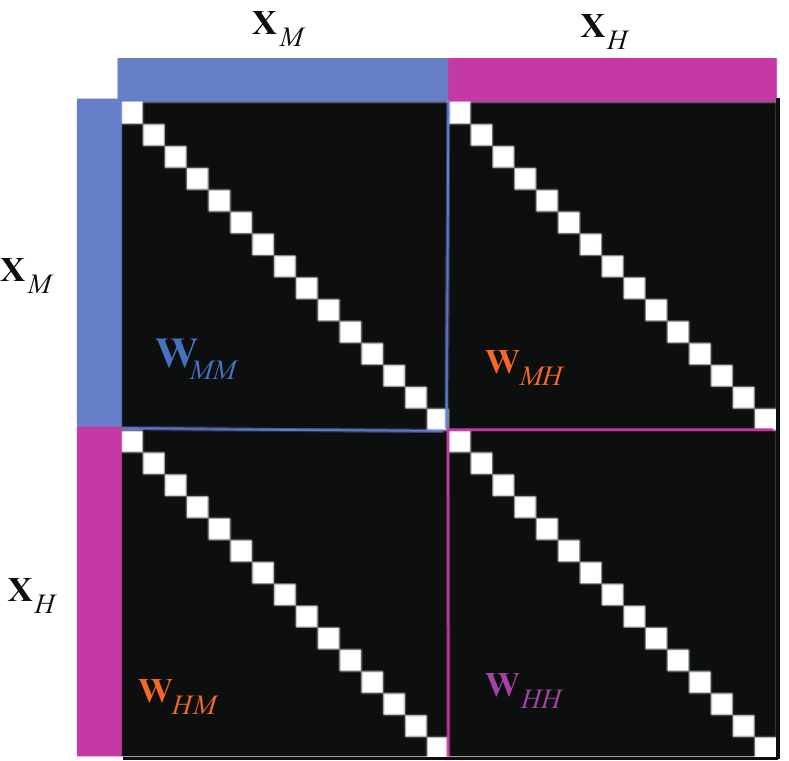}
		}
        \caption{An example to clarify the joint adjacency matrix.}
\label{fig:JointGraph}
\end{figure}

In order to solve (\ref{eq8}) effectively with ADMM, we consider an equivalent form by introducing auxiliary variables $\mathbf{J}$ and $\mathbf{G}$ to replace $\mathbf{\Theta}\widetilde{\mathbf{X}}$ and $\mathbf{\Theta}$, respectively.
\begin{equation}
\label{eq9}
\mathop{\min}_{\mathbf{\Theta},\mathbf{J},\mathbf{G}}\left\{
\begin{aligned}
&\frac{1}{2}\norm{\widetilde{\mathbf{Y}}-\mathbf{P}\mathbf{J}}_{\F}^{2}+\frac{\beta}{2}\tr(\mathbf{\Theta}\widetilde{\mathbf{X}}\mathbf{L}(\mathbf{\Theta}\widetilde{\mathbf{X}})^{\T})\\
       &\mathrm{s.t.} \quad \mathbf{J}=\mathbf{\Theta}\widetilde{\mathbf{X}}, \; \mathbf{G}=\mathbf{\Theta}, \; \mathbf{G}\mathbf{G}^{\T}=\mathbf{I}
\end{aligned}
\right\}.
\end{equation}
The augmented Lagrangian version of Eq. (\ref{eq9}) is
\begin{equation}
\label{eq10}
\begin{aligned}
       &\hspace{-0.7cm}\mathscr{L}_{C}\left(\mathbf{\Theta}, \mathbf{J}, \mathbf{G}, \mathbf{\Lambda}_{1}, \mathbf{\Lambda}_{2} \right)\\
       =&\frac{1}{2}\norm{\widetilde{\mathbf{Y}}-\mathbf{P}\mathbf{J}}_{\F}^{2}+\frac{\beta}{2}\tr(\mathbf{\Theta}\widetilde{\mathbf{X}}\mathbf{L}(\mathbf{\Theta}\widetilde{\mathbf{X}})^{\T})+\mathbf{\Lambda}_{1}^{\T}(\mathbf{J}-\mathbf{\Theta}\widetilde{\mathbf{X}})\\
       +&\mathbf{\Lambda}_{2}^{\T}(\mathbf{G}-\mathbf{\Theta})+\frac{\mu}{2}\norm{\mathbf{J}-\mathbf{\Theta}\widetilde{\mathbf{X}}}_{\F}^{2}+\frac{\mu}{2}\norm{\mathbf{G}-\mathbf{\Theta}}_{\F}^{2}\\
       &\mathrm{s.t.} \quad \mathbf{G}\mathbf{G}^{\T}=\mathbf{I},
\end{aligned}
\end{equation}
where $\mathbf{{\Lambda}_{1}}$ and $\mathbf{{\Lambda}_{2}}$ are Lagrange multipliers and $\mu$ is the penalty parameter. \textbf{Algorithm~2} summarizes the specific procedures for solving the problem (\ref{eq9}), and the solution to each subproblem is detailed in Appendix A.

Finally, we repeat these optimization procedures until a stopping criterion is satisfied.
\subsection{Convergence Analysis}
\begin{figure}[!t]
	  \centering
		\subfigure[University of Houston HS-MS Datasets]{
			\includegraphics[width=4.1cm,height=3cm]{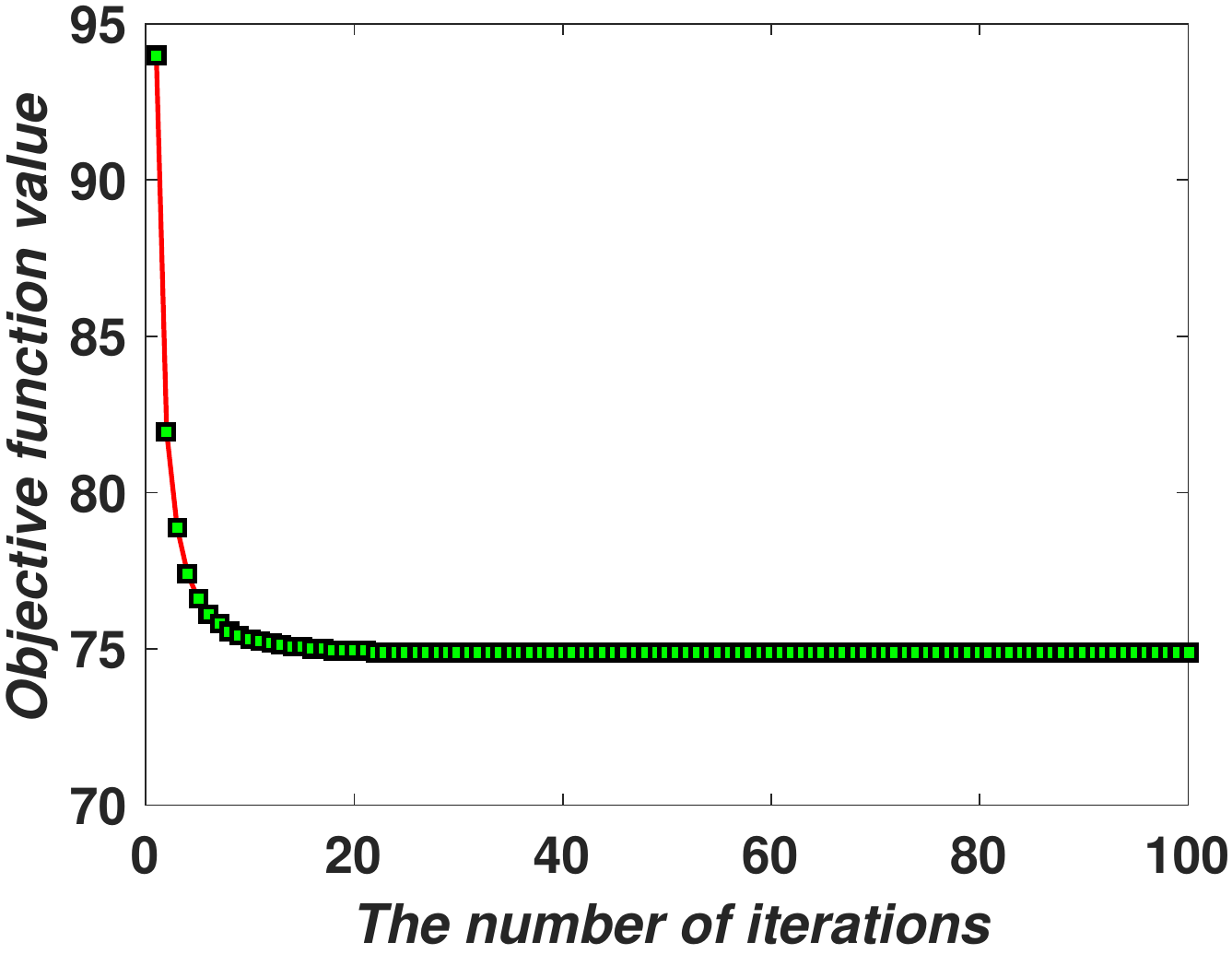}
            \label{fig:Convergence_HU}
		}
		\subfigure[Chikusei HS-MS Datasets]{
			\includegraphics[width=4.1cm,height=3cm]{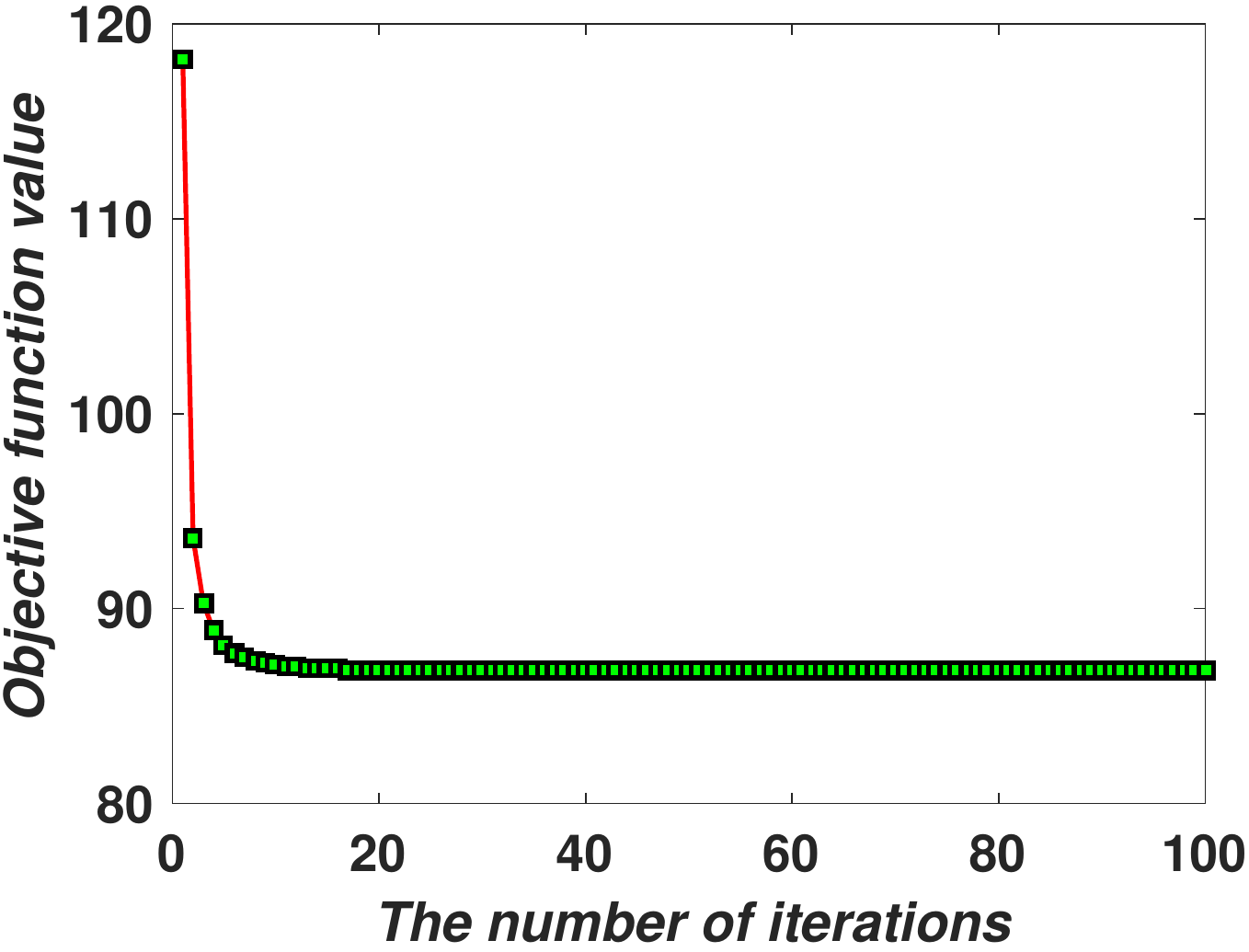}
            \label{fig:Convergence_CH}
		}
         \caption{Convergence analysis of CoSpace are experimentally performed on the two HS-MS datasets.}
\label{fig:Convergence}
\end{figure}

The iterative alternating strategy used in \textbf{Algorithm~1} and \textbf{Algorithm~2} is a block coordinate descent (BCD), whose convergence is theoretically guaranteed as long as each subproblem of Eq. (\ref{eq2}) is strictly convex, which can be exactly minimized \cite{bertsekas1999nonlinear}. Moreover, we experimentally display an illustration to clarify the convergence of CoSpace on both HS-MS datasets, where the objective function value is recorded in each iteration (see Fig. \ref{fig:Convergence}).
\section{Experiments}
In this section, we quantitatively and qualitatively evaluate the performance of the proposed method on two HS-MS datasets taken over the University of Houston and Chikusei. To validate the transferability of learned features by our CoSpace method, classification is explored as a potential application. Therefore, three different classifiers, namely the nearest neighbor (NN) based on the Euclidean distance, linear support vector machines (LSVM), and canonical correlation forest (CCF) \cite{CCF2015} are selected for this task. As a variant of random forest \cite{liaw2002classification}, CCF has shown its effectiveness in various tasks \cite{7580572,yokoya2018open,hong2018joint} owing to supervised feature extraction via canonical correlation analysis when constructing each decision tree. Furthermore, We compare the proposed method (CoSpace) with several classical approaches, which are suitable for the cross-modal feature learning task, including principle component analysis (PCA) based on joint dimensionality reduction (P-JDR for short) \cite{martinez2001pca}, locality preserving projection (LPP) based on unsupervised manifold alignment (L-USMA for short) \cite{he2004LPP}, and LPP-based supervised manifold alignment (L-SMA) \cite{wang2009general} as well as the original MS (Baseline).
\begin{figure*}[!t]
	  \centering
		\subfigure[University of Houston HS-MS Datasets]{
			\includegraphics[width=8cm,height=6cm]{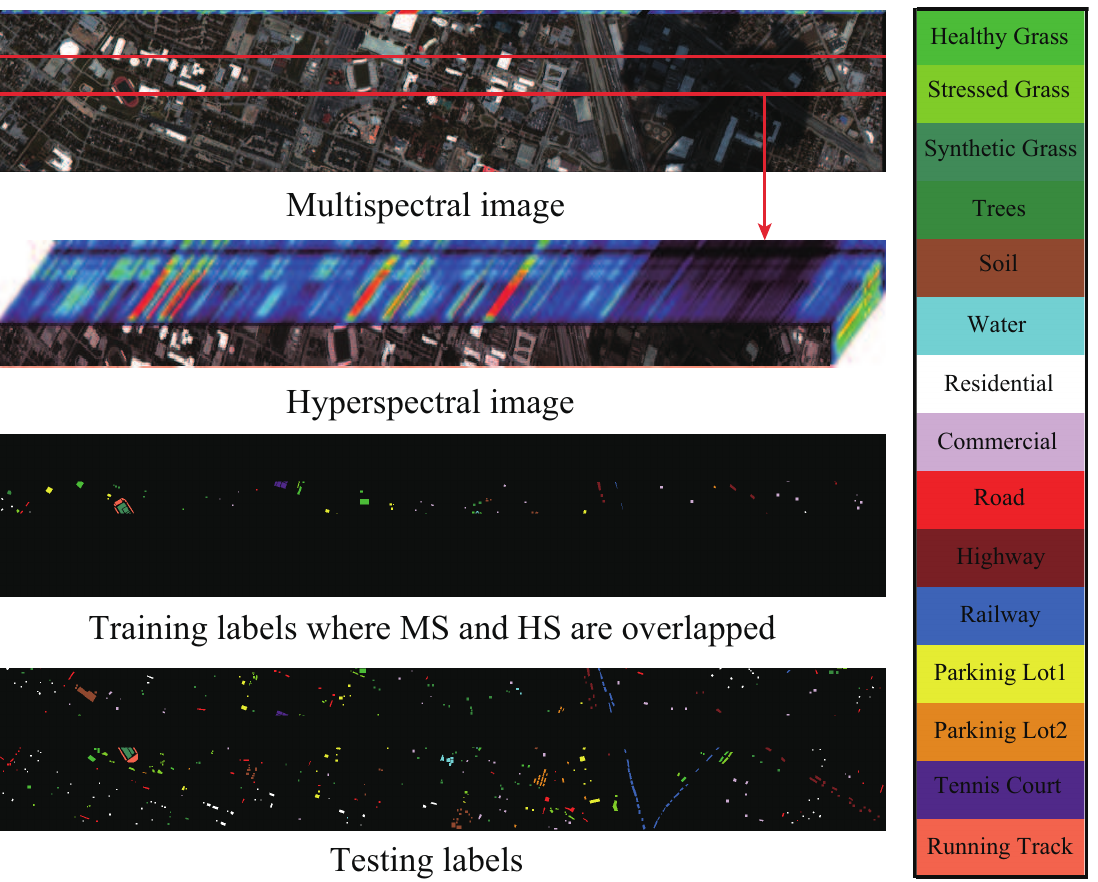}
            \label{fig:Data_H}
		}
		\subfigure[Chikusei HS-MS Datasets]{
			\includegraphics[width=8cm,height=6cm]{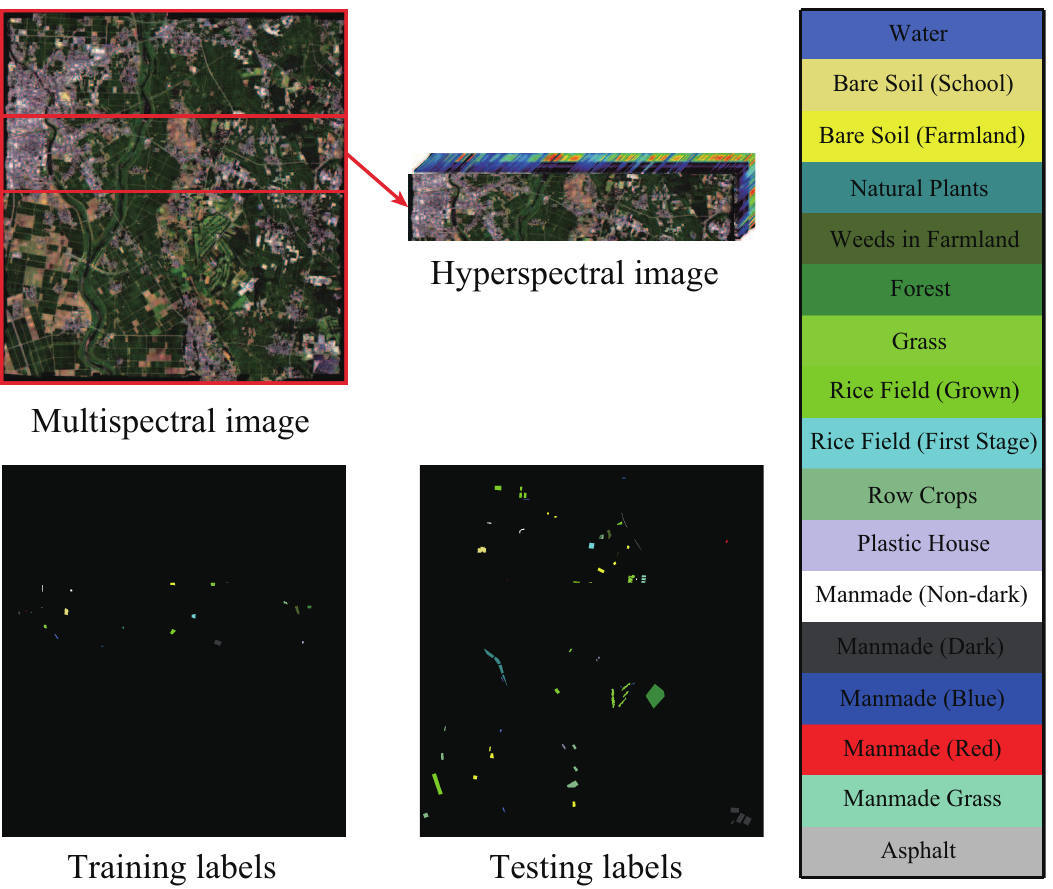}
            \label{fig:Data_CH}
		}
       \caption{The multispectral image and its corresponding hyperspectral image that partially covers the same area, as well as training and testing labels, for University of Houston dataset (a) and Chikusei Dataset (b), respectively.}
\label{fig:SceneImage}
\end{figure*}
\begin{table*}[!t]
\begin{minipage}{0.49\linewidth}
\footnotesize
\centering
\caption{The number of training and testing samples for the University of Houston MS-HS dataset.}
\smallskip
\begin{tabular}{cccc}
\hlinew{1.5pt}
Class No.&Class Name&Training&Testing\\
\hline \hline 1&Healthy Grass&537&699\\
 2&Stressed Grass&61&1154\\
 3&Synthetic Grass&340&357\\
 4&Tree&209&1035\\
 5&Soil&74&1168\\
 6&Water&22&303\\
 7&Residential&52&1203\\
 8&Commercial&320&924\\
 9&Road&76&1149\\
 10&Highway&279&948\\
 11&Railway&33&1185\\
 12&Parking Lot1&329&904\\
 13&Parking Lot2&20&449\\
 14&Tennis Court&266&162\\
 15&Running Track&279&381\\
\hline \hline &Total&2897&12021\\
\hlinew{1.5pt}
\end{tabular}
\label{Table:H}
\end{minipage}\begin{minipage}{0.49\linewidth}
\footnotesize
\centering
\caption{The number of training and testing samples for the Chikusei MS-HS dataset.}
\smallskip
\begin{tabular}{cccc}
\hlinew{1.5pt}
 Class No.&Class Name&Training&Testing\\
\hline \hline 1&Water&301&858\\
 2&Bare Soil (School)&992&1867\\
 3&Bare Soil (Farmland)&455&4397\\
 4&Natural Plants&150&4272\\
 5&Weeds in Farmland&928&1108\\
 6&Forest&486&11904\\
 7&Grass&989&5526\\
 8&Rice Field (Grown)&813&8816\\
 9&Rice Field (First Stage)&667&1268\\
 10&Row Crops&377&5961\\
 11&Plastic House&165&475\\
 12&Manmade (Non-dark)&170&568\\
 13&Manmade (Dark)&1291&6373\\
 14&Manmade (Blue)&111&431\\
 15&Manmade (Red)&35&187\\
 16&Manmade Grass&21&1019\\
 17&Asphalt&384&417\\
\hline \hline &Total&8335&55447\\
\hlinew{1.5pt}
\end{tabular}
\label{Table:CH}
\end{minipage}
\label{Table:Training and Testing}
\end{table*}
\subsection{University of Houston HS-MS Datasets}
\subsubsection{Data Description}
The HS data was acquired by the ITRES CASI-1500 sensor over an urban area around the campus of the University of Houston, Houston, USA, which was provided in the $2013$ IEEE GRSS data fusion contest \cite{debes2014hyperspectral}. The image consists of $349\times1905$ pixels with 144 spectral bands in the wavelength from 364 nm to 1046 nm with spectral resolution of 10 nm at a ground sampling distance (GDS) of 2.5 m.  Spectral simulation is performed to generate the MS image by degrading the full HS image in the spectral domain using the MS spectral response functions (SRFs) of Sentinel-2 as filters (for more details refer to \cite{yokoya2017hyperspectral}). Following this, the MS data with dimensions of $349\times1905\times10$ is generated. The MS image and the corresponding partial HS image over the University of Houston scene are shown in Fig. \ref{fig:Data_H}.
\begin{figure*}[!t]
	  \centering
		\subfigure{
			\includegraphics[width=0.95\textwidth]{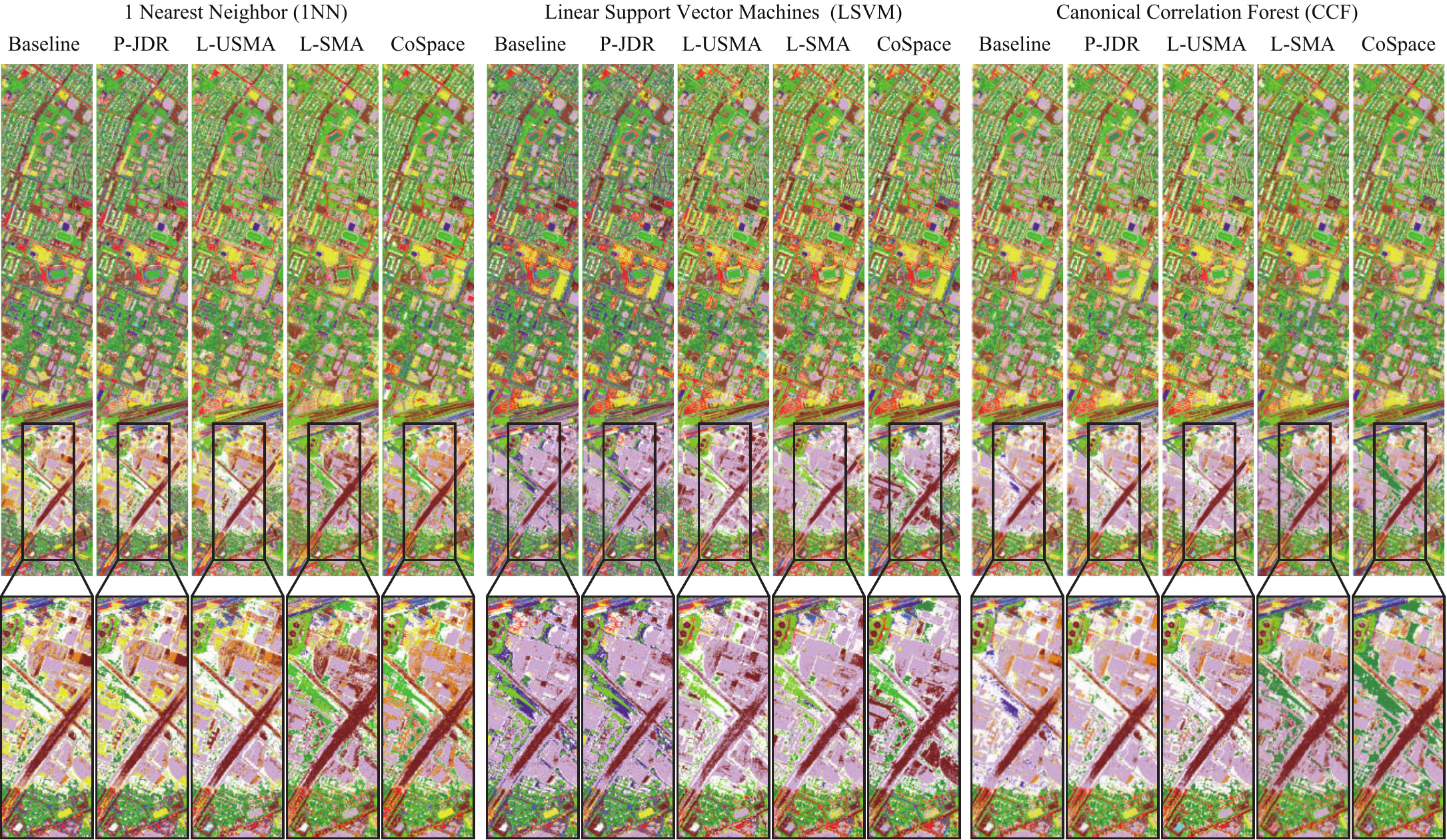}
		}
        \caption{Classification maps and corresponding highlighted sub-areas of the different algorithms obtained using three kinds of classifiers on the University of Houston dataset.}
\label{fig:ClassificationMap_H}
\end{figure*}
\begin{table*}[!t]
\scriptsize
\centering
\caption{Quantitative performance comparison with the different algorithms on the University of Houston data. The best one is shown in bold.}
\resizebox{\textwidth}{!}{
\begin{tabular}{c||c|c|c||c|c|c||c|c|c||c|c|c||c|c|c}
\hlinew{1.5pt}Algorithm&\multicolumn{3}{c||}{Baseline (\%) }&\multicolumn{3}{c||}{P-JDR (\%) }&\multicolumn{3}{c||}{L-USMA (\%) }&\multicolumn{3}{c||}{L-SMA (\%) }&\multicolumn{3}{c}{CoSpace (\%) }\\
\hline \hline Parameter&\multicolumn{3}{c||}{/}&\multicolumn{3}{c||}{$d=30$}&\multicolumn{3}{c||}{$(k,\sigma,d)=(10,1,20)$}&\multicolumn{3}{c||}{$d=30$}&\multicolumn{3}{c}{$(\alpha,\beta,d)=(0.01,0.01,30)$}\\
\hline \hline Classifier& 1NN & LSVM & CCF & 1NN & LSVM & CCF& 1NN & LSVM & CCF& 1NN & LSVM & CCF& 1NN & LSVM & CCF\\
\hline \hline OA&61.70&62.12&68.21&66.37&64.31&67.13&66.33&65.54&68.41&65.04&68.01&69.59&68.82&69.38&\textbf{72.17}\\
              AA&65.57&65.97&70.47&69.23&67.50&69.64&69.37&68.81&70.84&68.15&70.50&71.02&71.29&71.69&\textbf{73.56}\\
              $\kappa$&0.5842&0.5889&0.6543&0.6345&0.6118&0.6430&0.6342&0.6251&0.6565&0.6202&0.6520&0.6695&0.6613&0.6672&\textbf{0.6975}\\
\hline \hline Class1&69.10&76.39&67.95&73.39&71.82&70.24&80.40&78.68&67.67&81.69&75.25&68.53&\textbf{88.56}&75.54&69.96\\
              Class2&79.20&80.59&78.08&81.20&81.72&72.53&78.94&79.90&75.04&90.99&\textbf{97.57}&77.90&73.48&73.74&77.99\\
              Class3&\textbf{100.00}&\textbf{100.00}&\textbf{100.00}&\textbf{100.00}&\textbf{100.00}&\textbf{100.00}&\textbf{100.00}&\textbf{100.00}&\textbf{100.00}&\textbf{100.00}&\textbf{100.00}&\textbf{100.00}&\textbf{100.00}&\textbf{100.00}&\textbf{100.00}\\
              Class4&85.22&85.51&92.27&87.63&91.21&93.24&94.30&96.23&92.85&88.60&94.78&\textbf{98.74}&93.62&\textbf{98.74}&98.26\\
              Class5&98.89&99.06&99.40&98.72&98.72&99.32&\textbf{99.49}&99.40&98.89&99.40&98.97&99.14&98.97&99.40&99.40\\
              Class6&86.14&86.14&86.14&86.47&86.14&76.57&86.14&\textbf{86.47}&\textbf{86.47}&\textbf{86.47}&\textbf{86.47}&70.96&85.15&85.48&85.15\\
              Class7&36.49&50.62&63.76&59.6&53.78&51.04&49.79&50.21&63.18&58.19&72.32&77.14&69.99&73.98&\textbf{80.05}\\
              Class8&50.76&56.49&56.06&59.42&59.42&59.09&54.76&\textbf{66.23}&56.82&55.74&62.01&62.23&58.77&63.53&62.01\\
              Class9&60.92&56.22&70.58&64.14&63.01&\textbf{72.32}&63.45&65.19&69.10&47.17&49.96&61.27&64.40&59.79&64.93\\
              Class10&40.93&45.36&45.25&44.41&49.16&45.36&44.09&53.90&47.47&49.47&58.12&52.32&45.15&\textbf{64.14}&57.70\\
              Class11&39.16&27.43&43.88&36.03&35.53&39.41&45.91&29.79&43.71&35.36&28.86&36.46&44.14&36.54&\textbf{47.26}\\
              Class12&41.37&31.64&56.08&51.55&25.66&\textbf{65.82}&45.8&29.76&62.06&65.04&35.84&62.5&50.22&46.79&62.72\\
              Class13&0.45&0.00&0.67&0.67&0.00&2.45&0.22&0.00&1.11&\textbf{4.68}&0.00&0.00&0.89&0.00&0.45\\
              Class14&97.53&97.53&98.77&98.15&98.77&99.38&99.38&98.77&\textbf{100.00}&97.53&\textbf{100.00}&\textbf{100.00}&98.15&\textbf{100.00}&99.38\\
              Class15&97.38&96.59&\textbf{98.16}&97.11&97.64&97.9&97.9&97.64&\textbf{98.16}&97.9&97.38&\textbf{98.16}&97.9&97.64&\textbf{98.16}\\
\hlinew{1.5pt}
\end{tabular}}
\label{tab:Houston}
\end{table*}
\subsubsection{Experimental Setup}
Initially, we re-distribute the training and testing samples, as shown in Fig. \ref{fig:Data_H} and more specifically listed in Table \ref{Table:H}, to meet our problem setting that there is a large amount of the MS data (complete low-quality data) together with a limited amount of the HS data (incomplete high-quality data).

For the performance assessment of the algorithms, we adopt three criteria to quantify experiential results as

1) \emph{Overall Accuracy} (OA): This index is defined by the ratio between the number of multispectral samples that are correctly classified and the number of corresponding test samples.

2) \emph{Average Accuracy} (AA): We collect the classification accuracy of each class and average them to achieve an AA-based evaluation.

3) \emph{Kappa Coefficient} ($\kappa$): It statistically measures the agreement between the final classification map and the ground-truth map. Generally speaking, $\kappa$ is more robust and convincing than a simple percent-based agreement calculation (e.g. OA and AA), since the agreement occurring by chance is fully considered.

Furthermore, we experimentally maximize the performance of the different algorithms by tuning their parameters, such as dimension ($d$), regularization parameters ($\alpha, \beta, \gamma$), etc., using 10-fold cross-validation on training data. For the dimension ($d$) which is a common parameter for all algorithms, they can be selected ranging from $10$ to $50$ at an interval of $10$. For the number of nearest neighbors ($k$) and the standard deviation of Gaussian kernel function ($\sigma$) in L-USMA, we select them in the range of $\{10, 20,...,50\}$ and $\{10^{-2}, 10^{-1}, \allowbreak 10^{0}, 10^{1}, 10^{2}\}$, respectively, and two regularization parameters ($\alpha, \beta$) in CoSpace are both chosen from $\{10^{-2}, 10^{-1}, \allowbreak 10^{0}, 10^{1}, 10^{2}\}$.
\subsubsection{Results and Analysis}
Fig.\ref{fig:ClassificationMap_H} shows the classification maps of compared algorithms using three different classifiers, while Table \ref{tab:Houston} details the quantitative assessment results under the optimal parameters determined by cross-validation. 

Overall, after absorbing partial HS information, those ASFL approaches are prone to obtain a better classification result, compared to the baseline (only multispectral data). P-JDR steadily outperforms the baseline, especially using 1NN and LSVM classifiers, although its classification accuracy using CCF is slightly lower than baseline's. By embedding local topological structure of data, L-USMA performs better than baseline, and even P-JDR, showing stable results for three kinds of classifiers. With a more discriminative supervised information, L-SMA obtains more competitive results by locally constructing LDA-like graph, whose performance is basically superior to that of the baseline, P-JDR, and L-USMA. Unlike L-SMA that only aligns different modalities on a common subspace, the proposed CoSpace learns a latent subspace by aligning different modalities but also bridges the learned subspace with label information, achieving a best classification accuracy. Compared to baseline, P-JDR, L-USMA, and L-SMA, CoSpace increases the OAs of $7.12\%$, $2.45\%$, $2.49\%$ , and $3.78\%$, respectively, with 1NN classifier, and $7.26\%$, $5.07\%$, $3.84\%$, and $1.37\%$, respectively, with LSVM classifier, as well as $3.96\%$, $5.04\%$, $3.75\%$, and $2.58\%$, respectively, with CCF classifier. Likewise, there are similar trends for the other indices of AA and $\kappa$, which indicates that CoSpace tends to learn semantically meaningful features.

We can also observe from Fig. \ref{fig:Data_H} and Table \ref{Table:H} that the training samples collected in a very limited area badly results in the data unbalance between different classes. For instance, the number of training samples in \emph{Health Grass} is dozens of times as much as that in \emph{Water}, \emph{Railway}, \emph{Residential}, \emph{Commercial} and \emph{Parking Lot2}. This might make the classifier impossible to be trained effectively, since more attentions are paid on those classes with large-size samples, and, contrariwise, the small-scale classes play relatively less and even nothing.
\begin{figure}[!t]
	  \centering
		\subfigure[Houston HS-MS Datasets]{
			\includegraphics[width=0.22\textwidth]{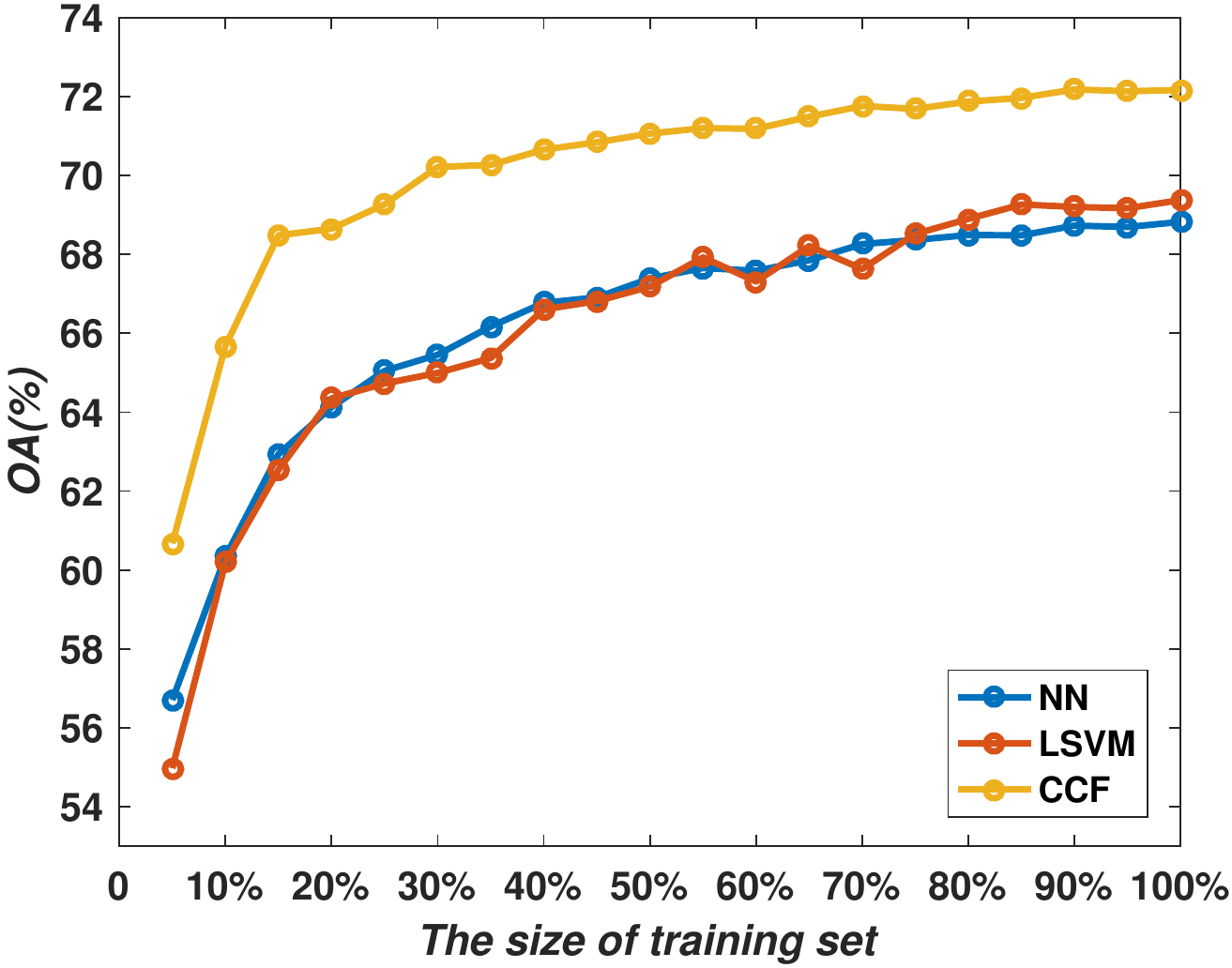}
            \label{fig:Sensi_TrainSize_H}
		}
		\subfigure[Chikusei HS-MS Datasets]{
			\includegraphics[width=0.22\textwidth]{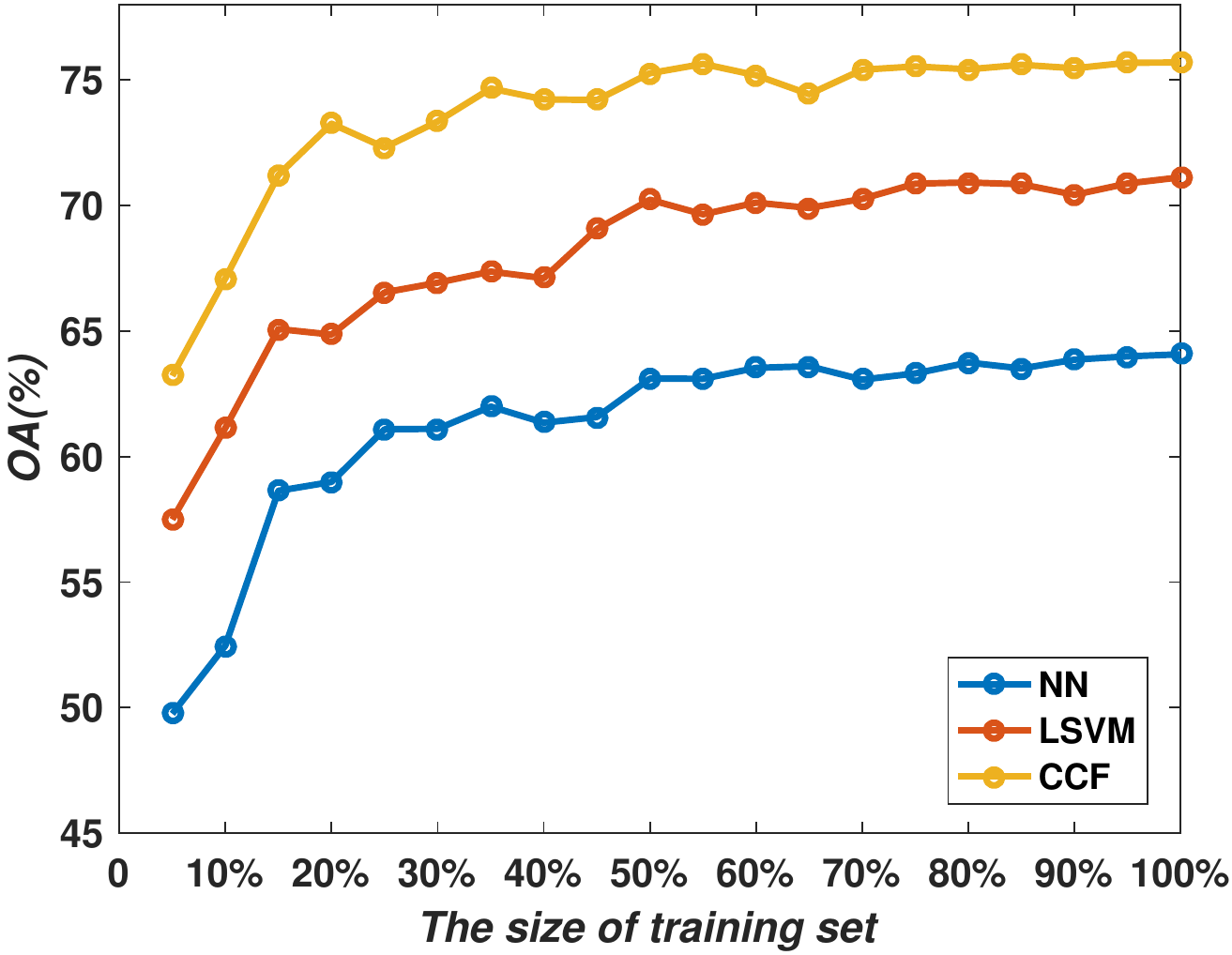}
            \label{fig:Sensi_TrainSize_CH}
		}
         \caption{Sensitivity analysis to the training set size using three different classifiers on the two MS-HS datasets.}
\end{figure}

A feasible solution to the problem is to enhance data representative ability by jointly feature learning from multi-modalities. For the performance evaluation of classifying those small-scale classes, e.g. \emph{Residential}, \emph{Commercial}, and \emph{Railway}, a direct evidence has been shown in Table \ref{tab:Houston} that those ASFL-based approaches (e.g. P-JDR, L-USMA, and L-SMA as well as CoSpace) obviously perform better on these small-scale samples than directly using original MS data (baseline). As expected, CoSpace dramatically outperforms the others, particularly on \emph{Residential} and \emph{Railway}. There is no denying, however, that CoSpace is superior to other algorithms to a larger extent, although it fails to effectively identify \emph{Parking Lot2} as same with others.

To visually highlight the classification differences for the different methods, we enlarge the classification maps of a sub-area overshadowed by the cloud, as shown in Fig. \ref{fig:ClassificationMap_H} where we can see that the methods with considering the hyperspectral information are able to generate the more discriminative features than the baseline, while the proposed CoSpace yields a better performance in identifying the materials in the shadow area, particularly for vegetation (e.g., \textit{Grass}), \textit{Residential} and \textit{Commercial} that are easily misclassified by the traditional methods.
\subsubsection{Sensitivity Analysis to the Training Set Size}
As the performance of the CoSpace largely depends on the number of training samples, it is, therefore, indispensable to investigate the sensitivity of the training set size. In detail, we conduct the classification using the CoSpace by fixing the test set and setting a series of new training sets randomly selected from the original training set with the different percentages ranging from 5\% to 100\% at a 5\% interval. As can be seen in Fig. \ref{fig:Sensi_TrainSize_H}, there is a similar trend in OAs using different classifiers, that is, the classification accuracy improves with the training set size, faster in the early, and later basically stabilized.
\begin{figure*}[!t]
	  \centering
		\subfigure{
			\includegraphics[width=0.9\textwidth]{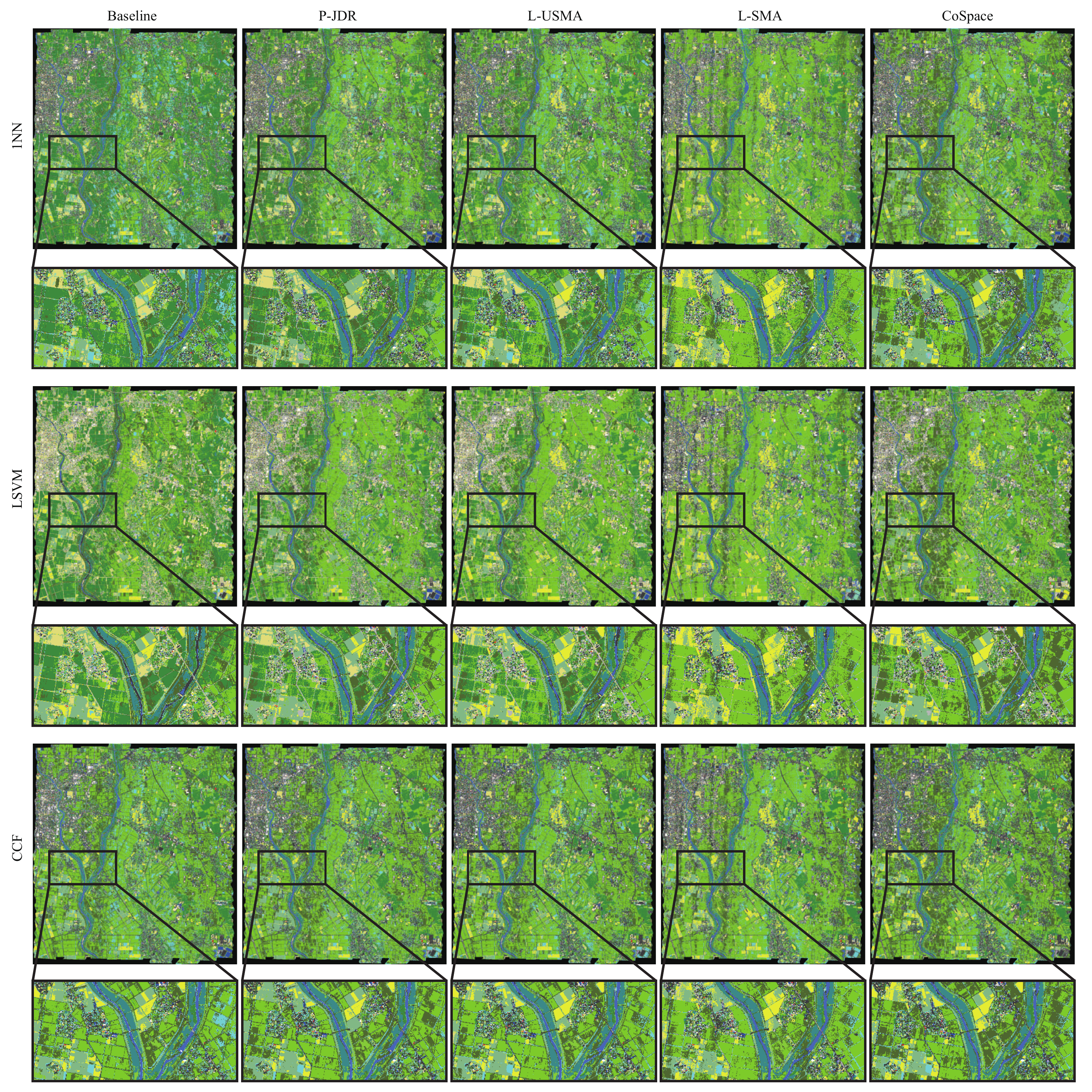}
		}
        \caption{Classification maps and corresponding highlighted sub-areas of the different algorithms obtained using three kinds of classifiers on the Chikusei dataset.}
\label{fig:ClassificationMap_CH}
\end{figure*}
\begin{table*}[!t]
\scriptsize
\centering
\caption{Quantitative performance comparison with the different algorithms on the Chikusei data. The best one is shown in bold.}
\resizebox{\textwidth}{!}{
\begin{tabular}{c||c|c|c||c|c|c||c|c|c||c|c|c||c|c|c}
\hlinew{1.5pt}Algorithm&\multicolumn{3}{c||}{Baseline (\%) }&\multicolumn{3}{c||}{P-JDR (\%) }&\multicolumn{3}{c||}{L-USMA (\%) }&\multicolumn{3}{c||}{L-SMA (\%) }&\multicolumn{3}{c}{CoSpace (\%) }\\
\hline \hline Parameter&\multicolumn{3}{c||}{/}&\multicolumn{3}{c||}{$d=30$}&\multicolumn{3}{c||}{$(k,\sigma,d)=(10,1,30)$}&\multicolumn{3}{c||}{$d=20$}&\multicolumn{3}{c}{$(\alpha,\beta,d)=(0.1,0.01,30)$}\\
\hline \hline Classifier& 1NN & LSVM & CCF & 1NN & LSVM & CCF& 1NN & LSVM & CCF& 1NN & LSVM & CCF& 1NN & LSVM & CCF\\
\hline \hline OA&55.99&60.20&71.11&60.45&68.19&71.87&61.48&67.31&72.33&62.44&67.90&71.53&64.07&71.12&\textbf{75.69}\\
              AA&62.39&69.42&70.40&64.95&71.71&71.02&64.95&70.06&71.49&64.52&70.79&66.47&68.05&7\textbf{3.96}&71.46\\
              $\kappa$&0.5084&0.5523&0.6761&0.5614&0.6414&0.6834&0.5715&0.6318&0.6893&0.5812&0.6391&0.6802&0.5995&0.6746&\textbf{0.7260}\\
\hline \hline Class1&79.49&78.21&80.54&97.2&98.25&81.93&98.48&\textbf{99.18}&81.00&80.89&98.72&82.52&80.19&92.54&79.25\\
              Class2&\textbf{95.02}&94.43&82.7&93.2&93.95&94.86&93.25&93.68&93.95&92.98&93.20&92.50&\textbf{95.02}&93.47&94.91\\
              Class3&11.37&23.54&50.06&14.40&16.03&24.13&31.07&39.37&56.74&61.81&62.57&55.31&80.10&\textbf{80.40}&77.71\\
              Class4&91.64&92.13&92.56&88.65&90.87&93.38&89.49&92.21&94.90&86.54&90.57&91.53&93.47&90.59&\textbf{96.23}\\
              Class5&87.00&\textbf{97.65}&94.68&67.78&70.94&76.26&63.90&34.57&79.87&25.36&28.43&16.06&74.28&83.94&66.52\\
              Class6&65.36&62.01&81.48&55.18&75.11&\textbf{87.77}&50.90&67.15&79.42&48.29&62.52&78.91&53.49&63.61&79.02\\
              Class7&99.26&99.67&99.93&91.95&98.43&\textbf{99.98}&91.02&97.00&99.71&95.46&96.87&97.79&88.74&97.74&99.75\\
              Class8&43.42&57.11&93.40&85.32&95.58&97.95&91.64&96.32&\textbf{99.29}&93.93&95.59&93.49&65.61&95.05&92.72\\
              Class9&99.92&\textbf{100.00}&\textbf{100.00}&98.34&98.66&99.53&98.82&99.45&99.92&99.21&99.53&99.13&\textbf{100.00}&98.66&99.76\\
              Class10&16.96&\textbf{24.81}&19.56&21.36&23.75&22.24&19.78&20.13&17.16&21.15&21.39&15.48&22.68&22.35&18.00\\
              Class11&2.11&0.00&2.11&0.63&1.05&\textbf{6.53}&0.00&0.00&3.37&0.00&0.00&0.00&0.00&0.00&0.00\\
              Class12&87.85&90.32&88.91&86.44&88.20&89.96&81.51&88.38&87.32&84.43&90.14&85.92&86.62&\textbf{90.32}&80.46\\
              Class13&33.08&33.11&33.09&33.08&33.11&33.11&32.36&33.01&33.11&25.47&32.61&56.25&33.11&33.11&\textbf{67.90}\\
              Class14&67.75&94.20&85.38&59.40&58.24&\textbf{99.77}&58.70&59.40&89.56&71.93&72.85&59.40&59.40&59.40&52.44\\
              Class15&97.33&\textbf{100.00}&\textbf{100.00}&\textbf{100.00}&\textbf{100.00}&\textbf{100.00}&96.26&96.79&\textbf{100.00}&94.12&93.58&\textbf{100.00}&\textbf{100.00}&93.58&97.86\\
              Class16&66.24&74.88&88.62&95.39&95.58&99.71&99.80&98.43&\textbf{100.00}&99.61&99.71&99.51&93.13&97.84&\textbf{100.00}\\
              Class17&16.79&58.03&3.84&15.83&\textbf{81.29}&0.24&7.19&76.26&0.00&15.59&65.23&7.91&30.94&64.75&0.00\\
\hlinew{1.5pt}
\end{tabular}}
\label{tab:Chikusei}
\end{table*}
\subsection{Chikusei HS-MS Datasets}
\subsubsection{Data Description}
The Headwall's Hyperspectral Visible and Near-Infrared series C (VNIR-C) imaging sensor acquired the airborne HS dataset over the agricultural and urban areas of Chikusei, Ibaraki, Japan, in 2014. This VNIR-C sensor collected 128 bands covering the wavelength range from 363 nm to 1018 nm with spectral resolution of 10 nm, and the scene consists of $2517 \times 2335$ pixels at GSD of 2.5 m.  The data set was made available to the scientific community recently, and more details regarding the data acquisition and processing can be found in \cite{yokoya2016airborne}. Similarly, the MS image with the size of $2517 \times 2335 \times 10$ was simulated by spectrally down-sampling the full HS image using the known SRFs of Sentinel-2. The generated MS image and the partial HS image over the Chikusei scene are shown in Fig. \ref{fig:Data_CH}.
\subsubsection{Experimental Setup}
Fig. \ref{fig:Data_CH} shows the latest training and testing labeling of the Chikusei dataset, which is quantified in Table \ref{Table:CH}. Three indices: OA, AA, and $\kappa$ introduced above are calculated to quantitatively assess the classification performance. Similarly to the case of the University of Houston dataset, the parameters for those given algorithms are determined by the 10-fold cross-validation on the training samples and the same range setting with those used for the University of Houston dataset is also conducted to the Chikusei dataset.
\subsubsection{Results and Analysis}
Similar to the University of Houston scene, we evaluate the performance for the Chikusei data both quantitatively and visually. Three classification indices with optimal parameters for different algorithms are summarized in Table \ref{tab:Chikusei}. For visual comparison, we give the corresponding classification maps in the full scene with those comparative algorithms under the different classifiers, as shown in Fig. \ref{fig:ClassificationMap_CH}.

As the classes in the Chikusei scene are more challenging classes and the distribution of training samples is inhomogeneous, directly using original MS data as input fails to identify certain materials, \emph{such as Forest}, \emph{Manmade (Dark)},\emph{ and Manmade (Grass)}, yielding a poor performance in OA, AA, and $\kappa$. Especially while using 1NN classifier, P-JDR and L-USMA, which belong to the unsupervised feature learning method, observably exceed basline in classification accuracy by $4.46\%$ and $5.49\%$, respectively. For LSVM and CCF classifiers, a similar trend is also demonstrated in Table \ref{tab:Chikusei}. Owing to the limited training samples and their distribution unbalance, the subspace projection learned by L-SMA  easily traps into over-fitting, despite of only having a weak performance improvement compared to these previously compared algorithms. By jointly performing subspace learning and classification, CoSpace not only aligns the different modalities in a latent common subspace, but also connects the subspace with label information formulated by training data. As a result, CoSpace obtains a higher classification accuracy than other algorithms, as listed in Table \ref{tab:Chikusei}. This might attribute to the learned common subspace, since the features projected in the subspace can absorb various properties from different modalities.

Similarly, we also make a visual comparison by giving a salient region in which the CoSpace's superiority in classifying complex and similar land-cover classes  is further shown as detailed in Fig. \ref{fig:ClassificationMap_CH}. Compared to other alignment-based methods, CoSpace is capable of better transferring HS information into MS data  by means of joint subspace learning and classification, yielding a more discriminative low-dimensional embedding. The learned features can recognize those classes of holding very similar features in MS data, such as \textit{Bare Soil (Farmland)} and \textit{Row Crops}, \textit{Weeds in Farmland} and \textit{Rice Field (Grown)}, more effectively. As shown in Fig. \ref{fig:ClassificationMap_CH}, CoSpace performs more reasonable and competitive classification results, that is, on one hand the \textit{Weeds in Farmland} and \textit{Rice Field (Grown)} are most likely to be coexisted in a scene; on the other hand, the \textit{Bare Soil (Farmland)} and \textit{Row Crops} are separated more correctly. This can be explained by a powerful transferability of HS information in the proposed CoSpace.
\subsubsection{Sensitivity Analysis to the Training Set Size}
Similar to the MS-HS Houston datasets, we apply the same investigating strategy and observe the trend of classification performance using CoSpace with different sizes of training sets on the MS-HS Chikusei datasets in Fig. \ref{fig:Sensi_TrainSize_CH}. There is a very substantial change in classification accuracy with the increase of the training set size ranging from 5\% to 40\% of total training samples, while the performance tends to be stable after the training set size is over 50\%.
\section{Conclusion}
The trade-off between MS and HS imaging in terms of observation ranges and spectral resolution motivates us to ponder whether HS data partially overlapping MS data can contribute to improving the classification performance of the whole MS imagery. For this purpose, we proposed CoSpace to achieve the property transferring in the different domains by learning a latent common subspace. Moreover, an effective joint strategy that simultaneously considers subspace learning and classification is embedded into the proposed method to tightly bridge the gap between the learned subspace and label information, leading to a more discriminative feature representation. The superior classification performance using CoSpace is demonstrated on two different datasets, compared to using other state-of-art methods. 

We performed transfer learning on homogeneous datasets in the considered MS-HS case in the sense that both data sources are optical images covering similar spectral ranges and thus the HS information can be transferred into the MS one linearly. The CoSpace's ability in handling heterogeneous data sources remains limited due to its linearized modeling. In the future work, we will develop a more general system by integrating some powerful and emerging nonlinear tools (e.g., deep learning) into our framework.

In addition, we just assumed to share the same land-cover classes across MS and HS images in this paper. In reality, the number of land-cover classes in the large-scale MS scene might be usually more than the one in the overlapped area of MS and HS images. This naturally motivates us to generalize our model in the future work.

\appendices
\section{Solution to problem (\ref{eq8}) with respect to $\Theta$}
The solution to problem (\ref{eq8}) can be transferred to equivalently solve the problem (\ref{eq10}) with ADMM. Considering the fact that the object function in Eq. (\ref{eq10})  is not convex with respect to all variables simultaneously, but it is a convex problem regarding the separate variable when other variables are fixed, therefore we successively minimize $\mathscr{L}_{C}$ with respect to $\mathbf{\Theta},\mathbf{J},\mathbf{G},\mathbf{\Lambda}_{1}, \mathbf{\Lambda}_{2}$ as follows:

{\it Optimization with respect to $\mathbf{\Theta}$:} The optimization problem for $\mathbf{\Theta}$ can be written as
\begin{equation}
\label{eq11}
\mathop{\min}_{\mathbf{\Theta}}\left\{
\begin{aligned}
&\frac{\beta}{2}\tr(\mathbf{\Theta}\widetilde{\mathbf{X}}\mathbf{L}(\mathbf{\Theta}\widetilde{\mathbf{X}})^{\T})+\mathbf{\Lambda}_{1}^{\T}(\mathbf{J}-\mathbf{\Theta}\widetilde{\mathbf{X}})\\
       +\mathbf{\Lambda}_{2}^{\T}(&\mathbf{G}-\mathbf{\Theta})+\frac{\mu}{2}\norm{\mathbf{J}-\mathbf{\Theta}\widetilde{\mathbf{X}}}_{\F}^{2}+\frac{\mu}{2}\norm{\mathbf{G}-\mathbf{\Theta}}_{\F}^{2}\}
\end{aligned}
\right\},
\end{equation}
which has a closed-form solution:
\begin{equation}
\label{eq12}
\begin{aligned}
       \mathbf{\Theta}\leftarrow&(\mu\mathbf{J}\widetilde{\mathbf{X}}^{\T}+\mathbf{\Lambda}_{1}\widetilde{\mathbf{X}}^{\T}+\mu\mathbf{G}+\mathbf{\Lambda}_{2})\\
            & \times(\mu\widetilde{\mathbf{X}}\widetilde{\mathbf{X}}^{\T}+\mu\mathbf{I}+\beta\widetilde{\mathbf{X}}\mathbf{L}\widetilde{\mathbf{X}}^{\T})^{-1}.
\end{aligned}
\end{equation}

{\it Optimization with respect to $\mathbf{J}$:} The variable $\mathbf{J}$ can be estimated by solving the following problem:
\begin{equation}
\label{eq13}
\mathop{\min}_{\mathbf{J}}\left\{
\begin{aligned}
\frac{1}{2}\norm{\widetilde{\mathbf{Y}}-\mathbf{P}\mathbf{J}}_{\F}^{2}+\mathbf{\Lambda}_{1}^{\T}(\mathbf{J}-\mathbf{\Theta}\widetilde{\mathbf{X}})+\frac{\mu}{2}\norm{\mathbf{J}-\mathbf{\Theta}\widetilde{\mathbf{X}}}_{\F}^{2}
\end{aligned}
\right\},
\end{equation}
its analytical solution is given by
\begin{equation}
\label{eq14}
\begin{aligned}
       \mathbf{J}\leftarrow(\mathbf{P}^{\T}\mathbf{P}+\mu\mathbf{I})^{-1}(\mathbf{P}^{\T}\widetilde{\mathbf{Y}}+\mu\mathbf{\Theta}\widetilde{\mathbf{X}}-\mathbf{\Lambda}_{1}).
\end{aligned}
\end{equation}

{\it Optimization with respect to $\mathbf{G}$:} For $\mathbf{G}$, the optimization problem with orthogonal constraint can be formulated as
\begin{equation}
\label{eq15}
\mathop{\min}_{\mathbf{G}}\left\{
\begin{aligned}
\mathbf{\Lambda}_{2}^{\T}(\mathbf{G}-\mathbf{\Theta})+\frac{\mu}{2}\norm{\mathbf{G}-\mathbf{\Theta}}_{\F}^{2}\}, \; \mathrm{s.t.} \; \mathbf{G}\mathbf{G}^{\T}=\mathbf{I}
\end{aligned}
\right\},
\end{equation}
which can be effectively solved using the strategy of splitting orthogonality constraints (SOC) \cite{lai2014splitting} in two steps:

the first step is to perform the \emph{SVD factorization}:
\begin{equation}
\label{eq16}
\begin{aligned}
      \left[\mathbf{U},\mathbf{S},\mathbf{V}\right]=\svd(\mathbf{\mathbf{\Theta}-\mathbf{\Lambda}_{2}/\mu}),
\end{aligned}
\end{equation}

the second step is to update $\mathbf{G}$ with satisfying orthogonal constraint:
\begin{equation}
\label{eq17}
\begin{aligned}
      \mathbf{G}\leftarrow\mathbf{U}\mathbf{I}_{n \times m}\mathbf{V}.
\end{aligned}
\end{equation}

{\it Lagrange multipliers ($\mathbf{\Lambda}_{1}$, $\mathbf{\Lambda}_{2}$) and penalty parameter ($\mu$) update:} Before stepping into the next iteration, Lagrange multipliers need to be updated by
\begin{equation}
\label{eq36}
\begin{aligned}
       \mathbf{\Lambda}_{1}\leftarrow\mathbf{\Lambda}_{1}+\mu (\mathbf{J}-\mathbf{\Theta}\widetilde{\mathbf{X}}),\quad \mathbf{\Lambda}_{2}\leftarrow\mathbf{\Lambda}_{2}+\mu (\mathbf{G}-\mathbf{\Theta}),
\end{aligned}
\end{equation}
and penalty parameter be updated by
\begin{equation}
\label{eq36}
\begin{aligned}
       \mu\leftarrow\ \min (\rho\mu, \mu_{\max}).
\end{aligned}
\end{equation}
\section*{Acknowledgments}

The authors would like to thank the Hyperspectral Image Analysis group and the NSF Funded Center for Airborne Laser Mapping (NCALM) at the University of Houston for providing the CASI University of Houston dataset. The authors would like to express their appreciation to Prof. D. Cai and Dr. C. Wang for providing MATLAB codes for LPP and manifold alignment algorithms.

\bibliographystyle{IEEEbib}
\bibliography{HDF_ref}

\begin{IEEEbiography}[{\includegraphics[width=1in,height=1.25in,clip,keepaspectratio]{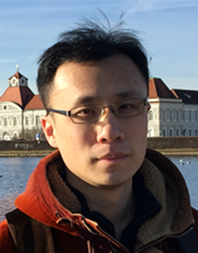}}]{Danfeng Hong}
(S'16) received the B.Sc. degree in computer science and technology from the
Neusoft College of Information, Northeastern University, China, in 2012, the M.Sc. degree in computer vision, Qingdao University, China, in 2015. He is currently pursuing his Ph.D. degree in Signal Processing in Earth Observation, Technical University of Munich (TUM), Munich Germany, and the Remote Sensing Technology Institute, German Aerospace Center (DLR), Oberpfaffenhofen, Germany.

His research interests include signal/image processing and analysis, pattern recognition, machine/deep learning and their applications in Earth Vision.
\end{IEEEbiography}
\vskip -2\baselineskip plus -1fil
\begin{IEEEbiography}[{\includegraphics[width=1in,height=1.25in,clip,keepaspectratio]{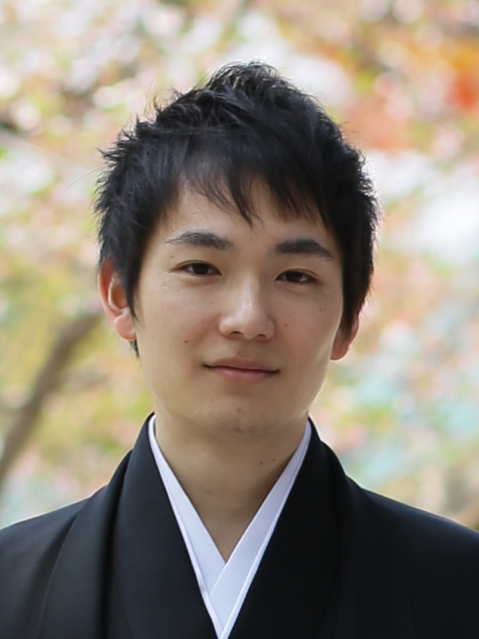}}]{Naoto Yokoya}
Yokoya (S'10–-M'13) received the M.Sc. and Ph.D. degrees in aerospace engineering from the University of Tokyo, Tokyo, Japan, in 2010 and 2013, respectively.

From 2012 to 2013, he was a Research Fellow with Japan Society for the Promotion of Science, Tokyo, Japan. From 2013 to 2017, he was an Assistant Professor with the University of Tokyo In 2015-2017, he was also an Alexander von Humboldt Research Fellow with the German Aerospace Center (DLR), Oberpfaffenhofen, and Technical University of Munich (TUM), Munich, Germany. Since 2018, he leads the Geoinformatics Unit at the RIKEN Center for Advanced Intelligence Project (AIP), RIKEN, Tokyo, Japan. His research interests include image analysis and data fusion in remote sensing.

Since 2017, he is a Co-chair of IEEE Geoscience and Remote Sensing Image Analysis and Data Fusion Technical Committee. In 2017, he won the Data Fusion Contest 2017 organized by the Image Analysis and Data Fusion Technical Committee of the IEEE Geoscience and Remote Sensing Society. His model was the most accurate among over 800 submissions.
\end{IEEEbiography}
\vskip -2\baselineskip plus -1fil
\begin{IEEEbiography}[{\includegraphics[width=1in,height=1.25in,clip,keepaspectratio]{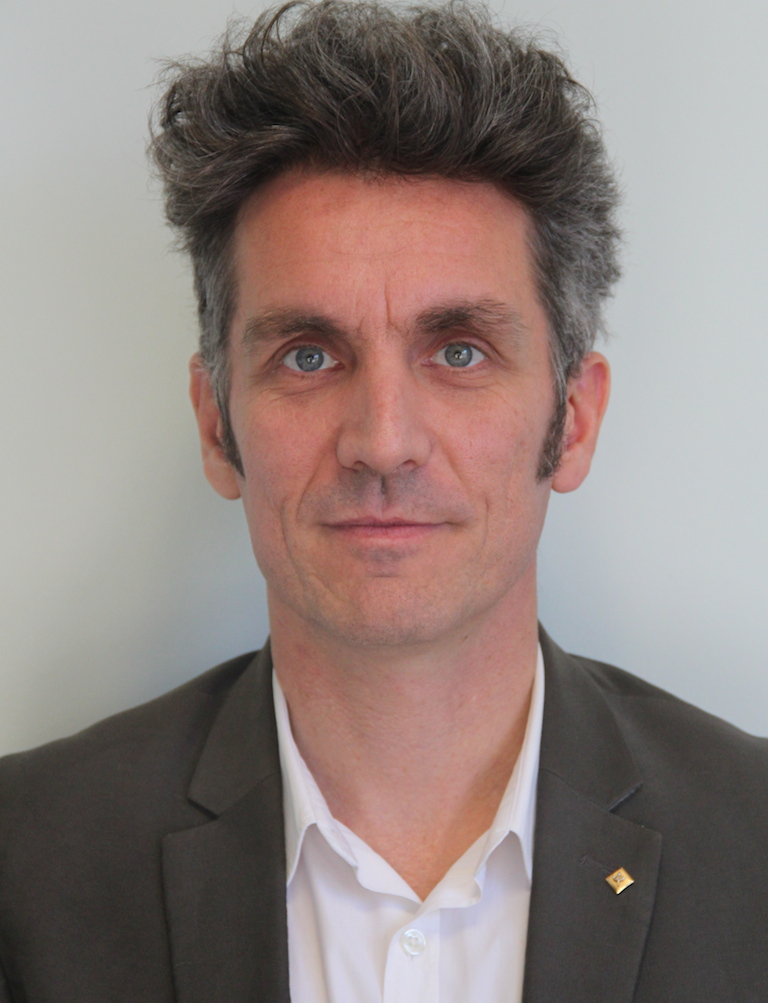}}]{Jocelyn Chanussot}
(M'04-–SM'04–-F'12) received the M.Sc. degree in electrical engineering from the Grenoble Institute of Technology (Grenoble INP), Grenoble, France, in 1995, and the Ph.D. degree from the Université de Savoie, Annecy, France, in 1998. In 1999, he was with the Geography Imagery Perception Laboratory for the Delegation Generale de l'Armement (DGA - French National Defense Department). Since 1999, he has been with Grenoble INP, where he is currently a Professor of signal and image processing. He is conducting his research at GIPSA-Lab. His research interests include image analysis, multicomponent image processing, nonlinear filtering, and data fusion in remote sensing. He has been a visiting scholar at Stanford University (USA), KTH (Sweden) and NUS (Singapore). Since 2013, he is an Adjunct Professor of the University of Iceland. In 2015-2017, he was a visiting professor at the University of California, Los Angeles (UCLA).

   Dr. Chanussot is the founding President of IEEE Geoscience and Remote Sensing French chapter (2007-2010) which received the 2010 IEEE GRSS Chapter Excellence Award. He was the co-recipient of the NORSIG 2006 Best Student Paper Award, the IEEE GRSS 2011 and 2015 Symposium Best Paper Award, the IEEE GRSS 2012 Transactions Prize Paper Award and the IEEE GRSS 2013 Highest Impact Paper Award. He was a member of the IEEE Geoscience and Remote Sensing Society AdCom (2009-2010), in charge of membership development. He was the General Chair of the first IEEE GRSS Workshop on Hyperspectral Image and Signal Processing, Evolution in Remote sensing (WHISPERS). He was the Chair (2009-2011) and Co-chair of the GRS Data Fusion Technical Committee (2005-2008). He was a member of the Machine Learning for Signal Processing Technical Committee of the IEEE Signal Processing Society (2006-2008) and the Program Chair of the IEEE International Workshop on Machine Learning for Signal Processing (2009). He was an Associate Editor for the IEEE Geoscience and Remote Sensing Letters (2005-2007) and for Pattern Recognition (2006-2008). He was the Editor-in-Chief of the IEEE Journal of Selected Topics in Applied Earth Observations and Remote Sensing (2011-2015). Since 2007, he is an Associate Editor for the IEEE Transactions on Geoscience and Remote Sensing, and since 2018, he is also an Associate Editor for the IEEE Transactions on Image Processing. In 2013, he was a Guest Editor for the Proceedings of the IEEE and in 2014 a Guest Editor for the IEEE Signal Processing Magazine. He is a Fellow of the IEEE and a member of the Institut Universitaire de France (2012-2017).
\end{IEEEbiography}

\begin{IEEEbiography}[{\includegraphics[width=1in,height=1.25in,clip,keepaspectratio]{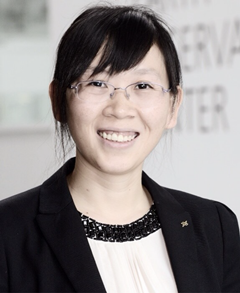}}]{Xiao Xiang Zhu}(S'10--M'12--SM'14) received the Master (M.Sc.) degree, her doctor of engineering (Dr.-Ing.) degree and her “Habilitation” in the field of signal processing from Technical University of Munich (TUM), Munich, Germany, in 2008, 2011 and 2013, respectively.

She is currently the Professor for Signal Processing in Earth Observation (www.sipeo.bgu.tum.de) at Technical University of Munich (TUM) and German Aerospace Center (DLR); the head of the department ``EO Data Science'' at DLR's Earth Observation Center; and the head of the Helmholtz Young Investigator Group ``SiPEO'' at DLR and TUM. Prof. Zhu was a guest scientist or visiting professor at the Italian National Research Council (CNR-IREA), Naples, Italy, Fudan University, Shanghai, China, the University  of Tokyo, Tokyo, Japan and University of California, Los Angeles, United States in 2009, 2014, 2015 and 2016, respectively. Her main research interests are
  remote sensing and Earth observation, signal processing, machine learning and data science, with a special application focus on global urban mapping.

  Dr. Zhu is a member of young academy (Junge Akademie/Junges Kolleg) at the Berlin-Brandenburg Academy of Sciences and Humanities and the German National  Academy of Sciences Leopoldina and the Bavarian Academy of Sciences and Humanities. She is an associate Editor of IEEE Transactions on Geoscience and Remote Sensing.
\end{IEEEbiography}

\end{document}